\documentclass[times,twocolumn,final]{elsarticle}
\usepackage{cag}
\usepackage{framed,multirow}

\usepackage{amssymb}
\usepackage{latexsym}

\usepackage{url}
\usepackage{xcolor}
\usepackage{color}
\usepackage{bm}
\definecolor{newcolor}{rgb}{.8,.349,.1}

\usepackage{hyperref}

\usepackage[switch,pagewise]{lineno} %Required by command \linenumbers below

\newcommand{\eg}{\emph{e.g.}}
\newcommand{\ie}{\emph{i.e.}}

\journal{Computers \& Graphics}

\begin{document}

\verso{Preprint Submitted for review}

\begin{frontmatter}

\title{SHREC’22 Track: Sketch-Based 3D Shape Retrieval in the Wild}%
% \tnotetext[tnote1]{Only capitalize first
% word and proper nouns in the title.}

\author[1]{Jie \snm{Qin}\corref{cor1}\fnref{fn1}}
\cortext[cor1]{Corresponding author:
%   Tel.: +0-000-000-0000;  
%   fax: +0-000-000-0000;
  }
\emailauthor{jie.qin@nuaa.edu.cn}{Jie Qin}
% \ead{example@email.com}
    
\author[2]{Shuaihang \snm{Yuan}\fnref{fn1}}
\fntext[fn1]{Track organizers.} 

\author[3]{Jiaxin \snm{Chen}\fnref{fn1}}

\author[4,5]{Boulbaba \snm{Ben Amor}\fnref{fn1}}

\author[2,6]{Yi \snm{Fang}\fnref{fn1}}

\author[7,9]{Nhat \snm{Hoang-Xuan}\fnref{fn2}}
\fntext[fn2]{Track participants.}

\author[7,9]{Chi-Bien \snm{Chu}\fnref{fn2}}

\author[7,9]{Khoi-Nguyen \snm{Nguyen-Ngoc}\fnref{fn2}}

\author[7,9]{Thien-Tri \snm{Cao}\fnref{fn2}}

\author[7,9]{Nhat-Khang \snm{Ngo}\fnref{fn2}}

\author[7,9]{Tuan-Luc \snm{Huynh}\fnref{fn2}}

\author[7,9]{Hai-Dang \snm{Nguyen}\fnref{fn2}}

\author[7,8,9]{Minh-Triet \snm{Tran}\fnref{fn2}}

\author[10]{Haoyang \snm{Luo}\fnref{fn2}}

\author[10]{Jianning \snm{Wang}\fnref{fn2}}

\author[10]{Zheng \snm{Zhang}\fnref{fn2}}

\author[11]{Zihao \snm{Xin}\fnref{fn2}}

\author[11]{Yang \snm{Wang}\fnref{fn2}}

\author[11]{Feng \snm{Wang}\fnref{fn2}}

\author[11]{Ying \snm{Tang}\fnref{fn2}}

\author[11]{Haiqin \snm{Chen}\fnref{fn2}}

\author[11]{Yan \snm{Wang}\fnref{fn2}}

\author[11]{Qunying \snm{Zhou}\fnref{fn2}}

\author[11]{Ji \snm{Zhang}\fnref{fn2}}

\author[11]{Hongyuan \snm{Wang}\corref{cor1}\fnref{fn2}}

\address[1]{Nanjing University of Aeronautics and Astronautics, Nanjing, China}
\address[2]{New York University, New York, USA}
\address[3]{Beihang University, Beijing, China}
\address[4]{IMT Nord Europe, Lille, France}
\address[5]{Inception Institute of Artificial Intelligence, Abu Dhabi, UAE}
\address[6]{New York University Abu Dhabi, Abu Dhabi, UAE}
\address[7]{University of Science, VNUHCM, Ho Chi Minh City, Vietnam}
\address[8]{John von Neumann Instittue, VNUHCM, Ho Chi Minh City, Vietnam}
\address[9]{Viet Nam National University, Ho Chi Minh City, Vietnam}
\address[10]{Harbin Institute of Technology, Shenzhen, China}
\address[11]{Changzhou University, Changzhou, China}

%\received{1 February 2017}
\received{\today}
%%%% Do not use the below for submitted manuscripts
%\finalform{28 March 2017}
%\accepted{2 April 2017}
%\availableonline{15 May 2017}
%\communicated{S. Sarkar}

\begin{abstract}
%%%
Sketch-based 3D shape retrieval (SBSR) is an important yet challenging task, which has drawn more and more attention in recent years. Existing approaches address the problem in a restricted setting, without appropriately simulating real application scenarios. To mimic the realistic setting, in this track, we adopt large-scale sketches drawn by amateurs of different levels of drawing skills, as well as a variety of 3D shapes including not only CAD models but also models scanned from real objects. We define two SBSR tasks and construct two benchmarks consisting of more than 46,000 CAD models, 1,700 realistic models, and 145,000 sketches in total. Four teams participated in this track and submitted 15 runs for the two tasks, evaluated by 7 commonly-adopted metrics. We hope that, the benchmarks, the comparative results, and the open-sourced evaluation code will foster future research in this direction among the 3D object retrieval community.
%%%%
\end{abstract}

\begin{keyword}
%% Keywords
\KWD Sketch-based 3D Shape Retrieval\sep Cross-modality Retrieval\sep Shape Retrieval in the Wild\sep Point Cloud Classification
\end{keyword}

\end{frontmatter}

%\linenumbers

\section{Introduction}
\label{intro}

Sketch-based 3D shape retrieval (SBSR) \cite{li2014comparison,chen2019deep,chen2018deep} has drawn a significant amount of attention, owing to the succinctness of free-hand sketches and the increasing demands from real applications. It is an intuitive yet challenging task due to the large discrepancy between the 2D and 3D modalities.

To foster the research on this important problem, several tracks focusing on related tasks have been held in the past SHREC challenges, such as \cite{li2013shrec,li2014shrec,li2016shrec,yuan2018shrec}. However, the datasets they adopted are not quite realistic, and thus cannot well simulate real application scenarios. To mimic the real-world scenario, the dataset is expected to meet the following requirements. First, there should exist a large domain gap between the two modalities, \ie, sketches and 3D shapes. However, current datasets unintentionally narrow this gap by using projection-based/multi-view representations for 3D shapes (\ie, a 3D shape is manually rendered into a set of 2D images). In this way, the large 2D-3D domain discrepancy is unnecessarily reduced to the 2D-2D one. Second, the data themselves from both modalities should be realistic, mimicking the real-world scenario. More specifically, we need a full variety of sketches per category as real users possess various drawing skills. As for 3D shapes, we need to frame 3D models with real-world settings more than create them artificially. However, human sketches on existing datasets tend to be semi-photorealistic drawn by experts and the number of sketches per category is quite limited; in the meantime, most current 3D datasets used in SBSR are composed of CAD models, losing certain details compared to 3D models scanned from real objects.

To circumvent the above limitations, this track proposes a more realistic and challenging setting for SBSR. On the one hand, we adopt highly abstract 2D sketches drawn by amateurs, and at the same time, bypass the projection-based representations for 3D shapes by directly adopting and representing 3D point cloud data. On the other hand, we adopt a full variety of free-hand sketches with various samples per category, as well as a collection of realistic point cloud data framed from indoor objects. Therefore, we name this track ‘sketch-based 3D shape retrieval in the wild’ (SBSRW). As stated above, the term ‘in the wild’ is reflected in two perspectives: 1) The domain gap between the two modalities is realistic as we adopt sketches of high abstraction levels and 3D point cloud data. 2) The data themselves mimic the real-world setting as we adopt a full variety of sketches and 3D point clouds captured from real objects.

\begin{figure}[t]
\centering 
\includegraphics[width=0.9\linewidth]{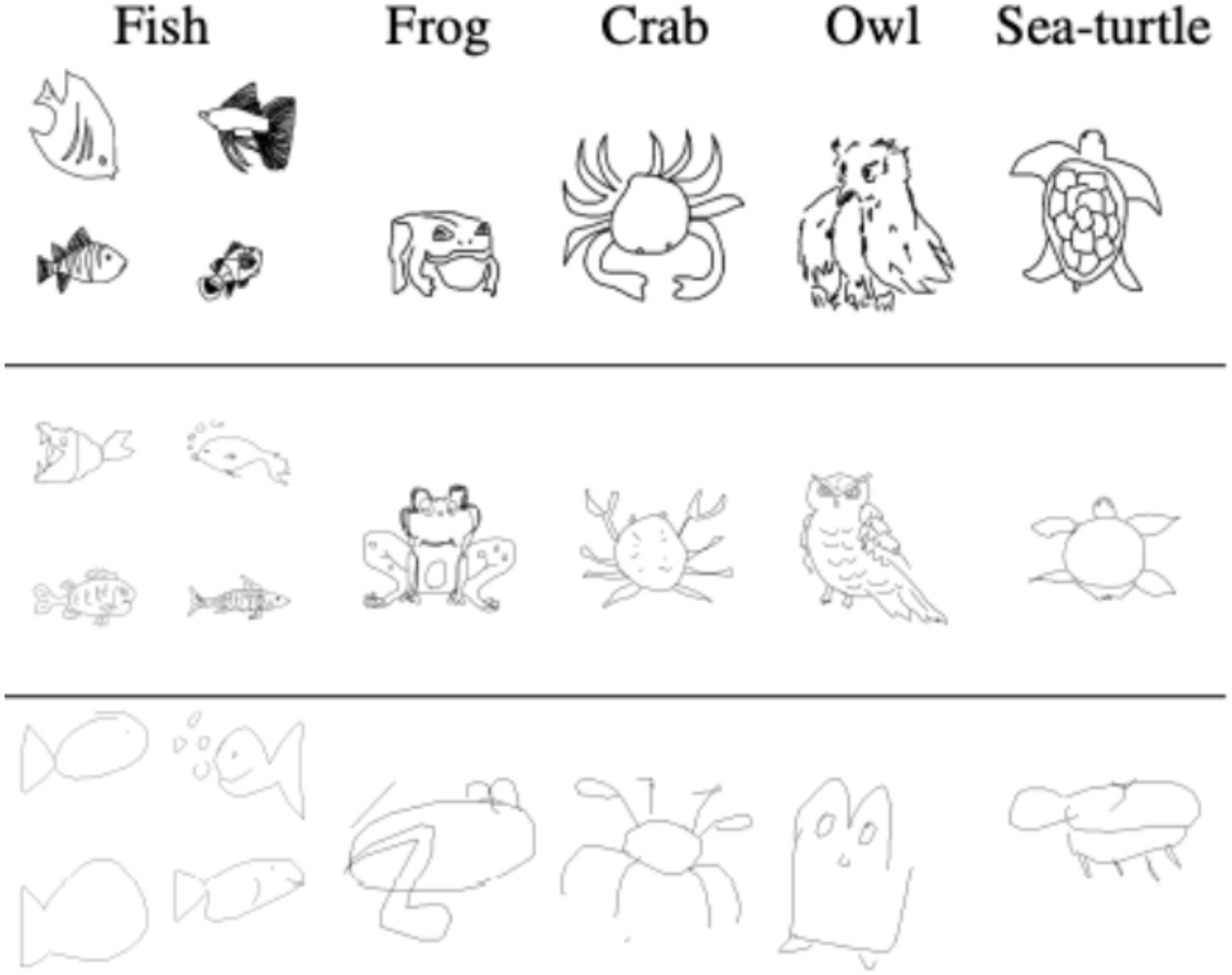}
\caption{Comparisons of some sketch samples between different benchmarks (images are from \cite{doodle}). The first, second, and third rows depict the sketches from Sketchy \cite{sketchy}, TU-Berlin \cite{tuberlin}, and QuickDraw \cite{quickdraw}, respectively.}
\label{fig:sketches}
\end{figure}
\begin{figure}[t]
\centering 
\includegraphics[width=\linewidth]{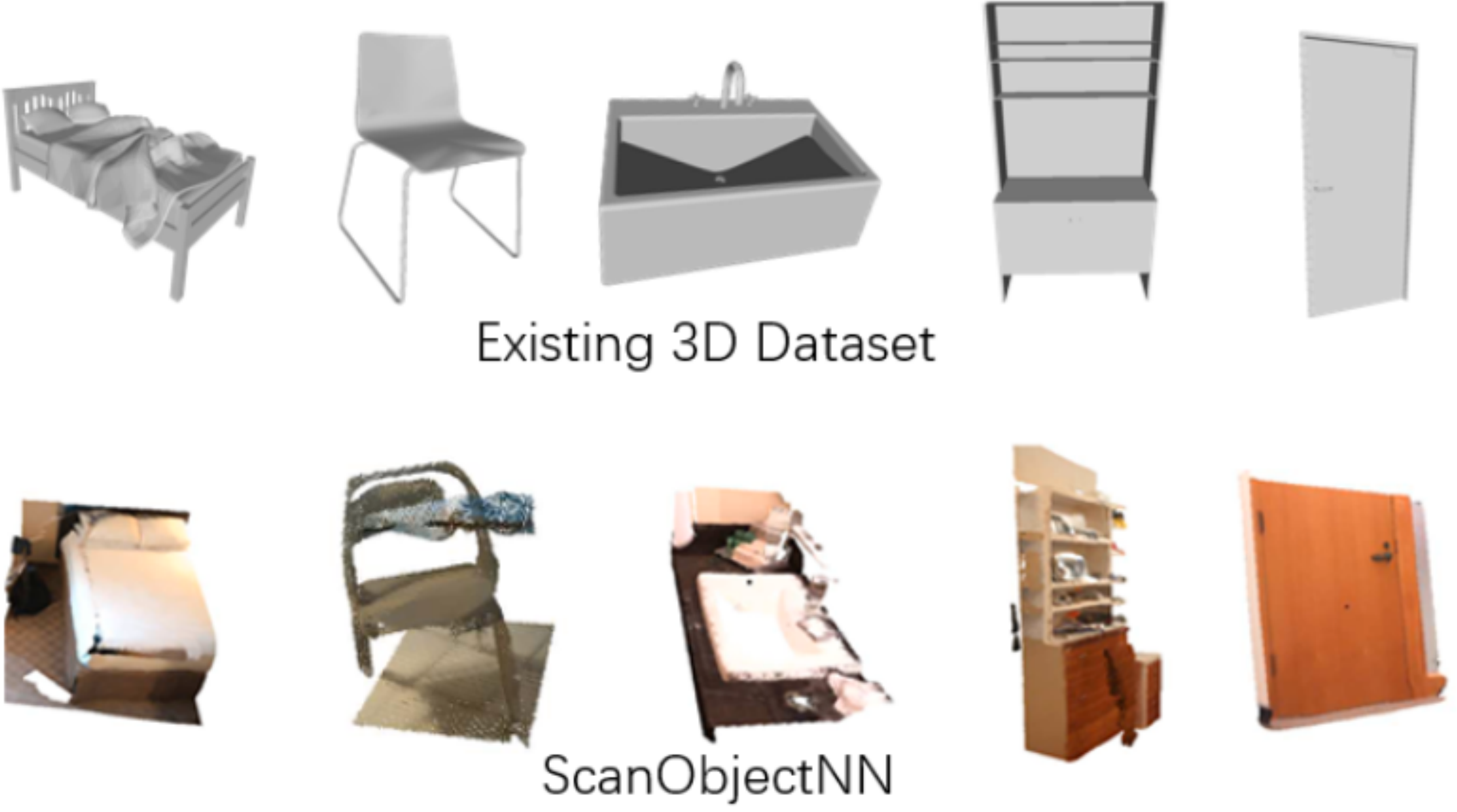}
\caption{Comparisons of synthetic 3D CAD models from existing datasets with 3D models from ScanObjectNN \cite{scanobjectnn} scanned in realistic scenarios.}
\label{fig:cadvswild}
\end{figure}

\begin{table*}[t]
\centering
\caption{Detailed statistics of different benchmarks for sketch-based 3D shape retrieval.}
\begin{tabular}{lcccccc}
\hline
\multirow{2}{*}{Datasets} & \multirow{2}{*}{\#Category} & \multicolumn{3}{c}{Sketches} & \multicolumn{2}{c}{Shapes} \\
\cline{3-7}
& & \#Training & \#Test & \#Avg/Class & \#Total & \#Avg/Class\\
\hline\hline
SHREC'13 & 90  &  4500 & 2700  & 80   & 1258  & 14   \\ 
SHREC'14 & 171 &  8550 & 5130  & 80   & 8987  & 53   \\ 
STC      & 44  & 96009 & 23992 & 2727 & 46612 & 1059 \\ 
STW      & 10  & 20135 & 5032  & 2517 & 1731  & 173  \\ 
\hline
\end{tabular}
\label{tab:stats}
\end{table*}

\begin{figure*}[t]
\begin{center}
   \includegraphics[width=0.95\linewidth]{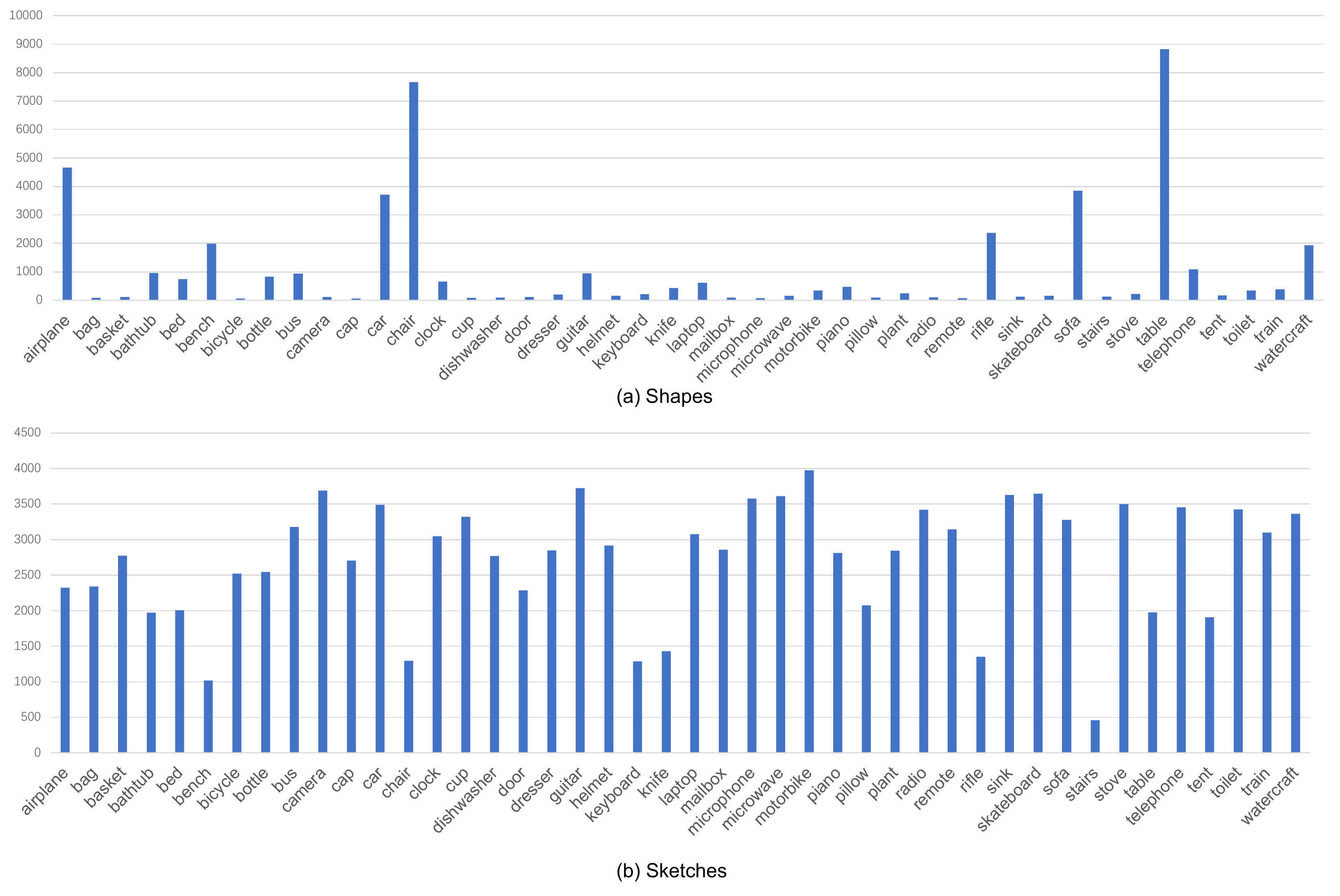}
\end{center}
\vspace{-5mm}
   \caption{Detailed statistics of the 44 classes on the STC benchmark w.r.t. Task 1.}
\label{fig:stc_stats}
\end{figure*}

\begin{figure*}[t]
\begin{center}
   \includegraphics[width=0.95\linewidth]{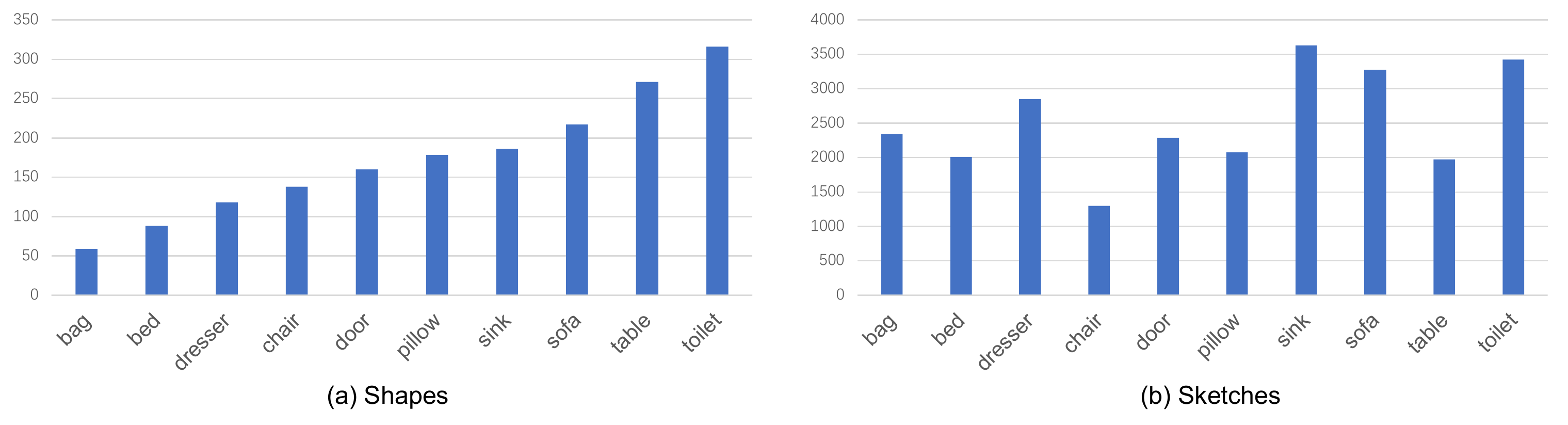}
\end{center}
\vspace{-5mm}
   \caption{Detailed statistics of the 10 classes on the STW benchmark w.r.t. Task 2.}
\label{fig:stw_stats}
\end{figure*}

\section{Benchmark Overview}
\label{data}

To fulfill our goal of SBSR in the wild, our benchmark takes advantage of four existing 2D/3D datasets, including a large-scale 2D sketch collection, \ie, QuickDraw \cite{quickdraw}, and three 3D point cloud datasets, \ie, ModelNet40 \cite{modelnet40}, ShapeNet \cite{shapenet}, and ScanObjectNN \cite{scanobjectnn}.

\textbf{QuickDraw} \cite{quickdraw} is a million-scale sketch dataset obtained via the online game ``QuickDraw'' created by Google. It contains over 50 million sketches belonging to 345 categories, collected from non-expert drawers around the world, with large variability derived from human abstraction. The large domain gap between non-expert drawers and photos is reflected on this dataset, which is not considered in previous sketch benchmarks. Some comparisons of the sketch samples between different benchmarks (including Sketchy \cite{sketchy} and TU-Berlin \cite{tuberlin}) are shown in Figure \ref{fig:sketches}.

\textbf{ModelNet40} \cite{modelnet40} is one of the most widely-used benchmark datasets in 3D shape analysis. The organizers of this dataset perform statistics analysis on the SUN dataset \cite{xiao2010sun} to find the most popular categories. Once those categories are determined, the corresponding CAD 3D shapes are collected by online searching. The collected shapes are then verified by human labors to ensure their correctness. As a result, 12,311 high-resolution 3D shapes are collected in the mesh format, and all 3D shapes are categorized into 40 classes.

\textbf{ShapeNet Core55} \cite{shapenet} is a subset of ShapeNet. It is a rich-annotated but challenging dataset for 3D shape analysis, especially for 3D shape retrieval and 3D part segmentation. There are 55 classes, among which around 57,000 3D meshes are distributed. 35,764, 5,133, and 10,265 3D shapes are selected for training, validation, and test, respectively.

\textbf{ScanObjectNN} \cite{scanobjectnn} includes 3D data in the real world represented by 3D point clouds. 3D shapes on this dataset are significantly different from those on the ShapeNet or the ModelNet40 datasets, due to the real-world background noise, occlusions, and \emph{etc}. It is composed of 2,902 3D point clouds from 15 categories. Three variants are proposed for different levels of difficulties. The vanilla version of the ScanObjectNN dataset contains only 3D point data cropped from the real-world scene scan using the ground truth bounding box. The second version contains the real-world indoor background information acquired during the scene scanning process. The last version contains the perturbed 3D shapes with background noise. In this track, we adopt the vanilla ScanObjectNN dataset containing clean 3D object point clouds. Some examples compared to the existing datasets are shown in Figure \ref{fig:cadvswild}.

Based on the existing 2D/3D datasets above, we construct our benchmark from two perspectives. First, we combine all the shapes from ModelNet40 and ShapeNet to form the large-scale collection of 3D CAD models. Due to some discrepancies between the categories of these two datasets, we finally obtain around 46,000 models from 44 classes in total. According to the models, we select an average of 2,700 sketches from the corresponding categories on QuickDraw. It is noteworthy that some sketches from QuickDraw are too abstract to be recognized even by human beings. So we manually remove those sketches as a preprocessing step. For brevity, we name this benchmark ``Sketch to CAD'' (STC). Second, all the scanned 3D objects from the vanilla ScanObjectNN dataset are adopted and the corresponding sketches are selected from QuickDraw in a similar manner as with the STC benchmark. Since several categories on ScanObjectNN are missing their counterparts on QuickDraw, we finally select around 1,700 realistic 3D models from 10 classes on ScanObjectNN and an average of 2,500 sketches per class from QuickDraw, constituting our ``Sketch to Wild'' (STW) benchmark. The detailed statistics of the STC and STW benchmarks can be found in Figures \ref{fig:stc_stats} and \ref{fig:stw_stats}, respectively. In addition, Table \ref{tab:stats} shows the comparison between the proposed two benchmarks and the existing SHREC'13 and SHREC'14 benchmarks. It is clearly observed that our proposed STC and STW datasets include significantly more examples for each category, making the SBSR task much more challenging and realistic.

\section{Evaluation}
\label{eval}
In accordance with the constructed STC and STW benchmarks, we proposed two tasks to evaluate the performance of different SBSR algorithms, \ie, sketch-based 3D CAD model retrieval and sketch-based realistic scanned model retrieval. Note that all 3D models in both tasks are provided in the form of point cloud data.

Specifically, in terms of the first task on the STC benchmark, we randomly select 80\% sketches from each class for training, and the remaining 20\% sketches per class are used for testing/query. Considering the large scale of 3D shapes, we randomly choose 20\% of the total 3D models as the target/gallery dataset to evaluate the retrieval performance. As for the second task on the STW benchmark, similar to the first task, we randomly select 80\% sketches from each class for training, and the remaining 20\% sketches per class are used for testing/query. Since the number of 3D models is not that large, all the 3D point clouds as a whole are utilized as the target/gallery dataset to evaluate the retrieval performance. Participants are asked to submit the results on the test sets.

\subsection{Evaluation Metric}
\label{sec:eval}
For a comprehensive evaluation of different algorithms, we employ the following widely-adopted performance metrics in SBSR, including nearest neighbor (NN), first tier (FT), second tier (ST), E-measure (E), discounted cumulated gain (DCG), mean average precision (mAP), and precision-recall (PR) curve. The source code to compute all the aforementioned metrics is provided. More specifically, during test, all the participants are required to submit the distance matrices computed based on the test data, where the $(i,j)$-th entry indicates the distance between the $i$-th sketch and the $j$-th shape. The smaller the distance is, the more similar the corresponding sketch and shape are.

\section{Participants}
\label{part}
A total of 7 teams registered for the SHREC'22 Track on Sketch-Based 3D Shape Retrieval in the Wild and eventually 4 of them submitted the final test results. For the first task, 6 rank list results (runs) for 3 different methods developed by 3 teams have been submitted. For the second task, 4 teams have developed 4 different methods, resulting in the submission of 9 rank list results (runs). The participants and their runs are listed as follows:
\begin{itemize}
    \item \textbf{Team A}: Thien-Tri Cao, Nhat-Khang Ngo, Tuan-Luc Huynh, Hai-Dang Nguyen, and Minh-Triet Tran from VNUHCM-University of Science submitted 1 run for Task 2 (Section \ref{sec:hcmus1}).
    \item \textbf{Team B}: Nhat Hoang-Xuan, Chi-Bien Chu, Khoi-Nguyen Nguyen-Ngoc, Hai-Dang Nguyen, and Minh-Triet Tran from VNUHCM-University of Science submitted 2 runs for Task 1 and 4 runs for Task 2, respectively (Section \ref{sec:hcmus2}).
    \item \textbf{Team C}: Haoyang Luo, Jianning Wang, and Zheng Zhang from Harbin Institute of Technology submitted 1 run for Task 1 and 1 run for Task 2, respectively (Section \ref{sec:hit}).
    \item \textbf{Team D}: Zihao Xin, Yang Wang, Feng Wang, Ying Tang, Haiqin Chen, Yan Wang, Qunying Zhou, Ji Zhang, and Hongyuan Wang from Changzhou University submitted 3 runs for task 1 and 3 runs for task 2, respectively (Section \ref{sec:cczu}).
\end{itemize}

\begin{figure*}[t]
\begin{center}
   \includegraphics[width=0.95\linewidth]{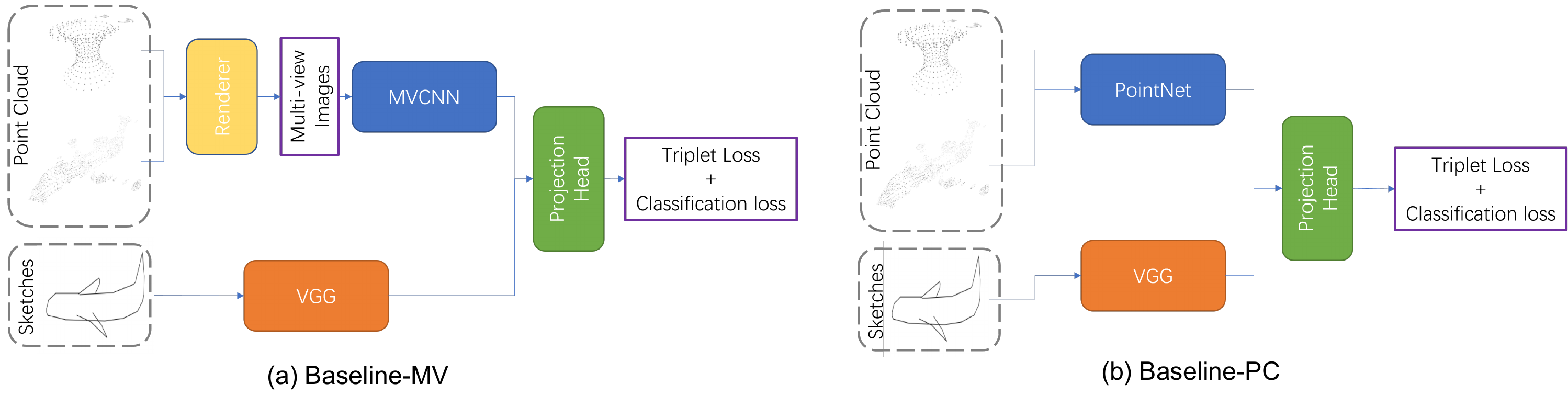}
\end{center}
\vspace{-5mm}
   \caption{Overall frameworks of (a) the multi-view baseline (Baseline-MV) and (b) the point cloud based baseline (Baseline-PC).}
\label{fig:baseline}
\end{figure*}

\section{Methods}
\label{method}

In this section, in line with our motivation, we first introduce two kinds of baseline methods (\ie, a projection-based/multi-view method and a point cloud based one). Then, we present the methods and describe the runs developed by the 4 teams in detail. Since the two tasks only differ in the data source (one is from CAD and the other from realistic scanner), the participated teams basically develop one general cross-modality retrieval framework, which is then trained based on different training sets.

\subsection{Baseline Methods}
Prior to introducing the methods provided by the participated teams, we first present two baseline methods. Specifically, we keep the sketch feature extraction head the same for two different baselines and vary the 3D model feature extraction head. For the 3D model feature extraction, the first baseline method renders a 3D model into multi-view images, based on which a multi-view method is used for 2D feature extraction. Differently, the second baseline method takes raw point cloud data as input and directly extracts 3D features without the additional rendering step.

\subsubsection{Multi-View Method}
\label{sec:baseline-mv}
As shown in Figure \ref{fig:baseline}(a), the multi-view baseline (\ie, Baseline-MV) contains three steps: 1) 3D model rendering, 2) multi-view feature extraction, and 3) sketch feature extraction.

\textbf{3D Model Rendering.}
We first render 3D models in the point cloud format to multi-view 2D images. For each 3D point cloud, we render $k$ images that correspond to the projections from different camera views. The rendered images are then fed to a feature learning module to learn multi-view features for the 3D point cloud. 

\textbf{Multi-View Feature Extraction.}
We follow multi-view convolutional neural networks (MVCNN) \cite{su15mvcnn} to extract multi-view based 3D shape features. As stated above, instead of using raw point clouds as input, the rendered 2D images are fed to the network. The multi-view feature extractor first generates 2D image descriptors for each view and then aggregates each individual descriptor by a max-pooling layer. The extracted multi-view shape features are then fed to multi-layer perceptrons (MLPs) to project the features to a common latent space.

\textbf{Sketch Feature Extraction.}
We employ the VGG \cite{vgg} network for sketch representation learning. Due to the high abstraction of the 2D sketches in this track, we pre-train the VGG network on the ImageNet and fine-tune the network parameters on our newly proposed datasets using the classification loss to better capture the semantic information. Similar to the shape feature extractor, we further adopt a projection head composed of MLPs to project the sketch features to a common latent space shared by the projected shape features.

\textbf{Objective Function.}
We train the multi-view baseline with three objective functions. Specifically, our model takes triplets as input. Given $N$ triplets $\{x_i^{sketch},x_i^{+},x_i^{-}\}$, where $x_i^{sketch}$ is a sketch image, $x_i^{+}$ represents a positive point cloud that belongs to the same category as the sketch image, and $x_i^{-}$ indicates a negative point cloud that is randomly selected from other categories, the first objective function measures the classification ability of the feature extractor, for which we use the cross entropy loss:
\begin{equation}
\centering
    \mathcal{L}_{sketch}=-\frac{1}{N} \sum_{i=1}^{N}y_{x_i^{sketch}} log(p_{x_i^{sketch}}),
    \label{eq:ce1}
\end{equation}
where $N$ is the total number of sketch samples, $y_{x_i^{sketch}}$ is the ground-truth label, and $p_{x_i^{sketch}}$ is the predicted softmax probability.

In addition to the loss term above, we also adopt the cross entropy loss to measure the quality of the point cloud feature extractor:
\begin{equation}
\centering
    \mathcal{L}_{shape}=-\frac{1}{N} \sum_{i=1}^{N}(y_{x_i^{+}} log(p_{x_i^{+}})+y_{x_i^{-}} log(p_{x_i^{-}})),
    \label{eq:ce2}
\end{equation}
where $N$ is the total number of 3D shapes, $y_{x_i^{+}}$ and $y_{x_i^{-}}$ are the ground-truth labels of positive and negative samples, respectively, and $p_{x_i^{+}}$ and $p_{x_i^{-}}$ are the predicted softmax probabilities for the positive and negative samples, respectively.

Finally, to capture the correspondence between 2D sketches and 3D shapes, we further adopt a soft-margin triplet loss as follows:
\begin{equation}
       \mathcal{L}_{tri}
       =\sum_{i=1}^{N} ln(1+exp(\|z_{x_i^{sketch}} - z_{x_i^{+}}\|^2-\|z^{x_i^{sketch}} - z_{x_i^{-}}\|^2)),
    \label{eq:tri}
\end{equation}
where $z_{x_i^{sketch}}$, $ z_{x_i^{+}}$ and $ z_{x_i^{-}}$ denote the projected sketch feature, positive multi-view 3D shape feature, and negative multi-view 3D shape feature, respectively.

\textbf{Implementation Details.}
We use the PyTorch3D library to render the 3D point clouds. We adopt 12 cameras uniformly distributed on a unit sphere where the 3D object is placed at the center of this sphere. In other words, we render each 3D shape into 12 2D images for 3D multi-view feature extraction. As stated above, we use the MVCNN and VGG networks to extract 3D point cloud features and sketch features, respectively. We use MLPs with three hidden layers to project 2D sketch and 3D shape features onto a common latent space. The Adam optimizer with a batch size of 32 is used to train the entire network. The momentum is set to 0.8, and the weight decay rate is set to $10^{-4}$. The learning rate starts at 0.001 and is then decreased by a factor of 3 every 40 epochs.

\subsubsection{Point Cloud Based Method}
As shown in Figure \ref{fig:baseline}(b), the point cloud based baseline (\ie, Baseline-PC) contains two steps: 1) point cloud feature extraction, and 2) sketch feature extraction.

\textbf{Point Cloud Feature Extraction.}
In contrast to the multi-view baseline, this baseline directly receives raw point clouds as input, avoiding the time-consuming rendering process. Specifically, we use PointNet \cite{pointnet} as the backbone network for 3D representation learning of the input point clouds. Concretely, PointNet utilizes multi-layer perceptrons (MLPs) to extract per-point feature signatures in the latent space and fuses signatures into high-level global representation by adopting a max-pooling layer. The extracted latent point cloud representations are further fed to a projection head to transform the point cloud features to a common latent space shared by the projected sketch features. We also employ MLPs as the projection head.

\textbf{Sketch Feature Extraction.}
Similar to the sketch feature extractor mentioned in the multi-view baseline, we use the pre-trained VGG network and fine-tune it on our newly proposed datasets to extract sketch features. The feature projection head is applied to the extracted sketch features for learning the common latent space.

\textbf{Objective Function.}
We train the point cloud based method using the same objective functions as the multi-view baseline, including two cross-entropy losses (\ie, Eqs. (\ref{eq:ce1}) and (\ref{eq:ce2})) and one triplet loss (\ie, Eq. \ref{eq:tri})), as described in Sec. \ref{sec:baseline-mv}.

\textbf{Implementation Details.}
The point could based method is trained end-to-end based on the aforementioned objective functions. We use an Adam optimizer with the momentum set to 0.7 and the weight decay rate set to 0.005. The initial learning rate is 0.001 and we decrease it by a factor of 10 every 60 epochs. During training, we use a batch size of 32 and sample 1024 points for each 3D point cloud.

\subsection{Team A: Distance-based Methods for Sketch-based 3D Shape Retrieval}
\label{sec:hcmus1}

\begin{figure}[b]
\centering 
\includegraphics[width =\linewidth]{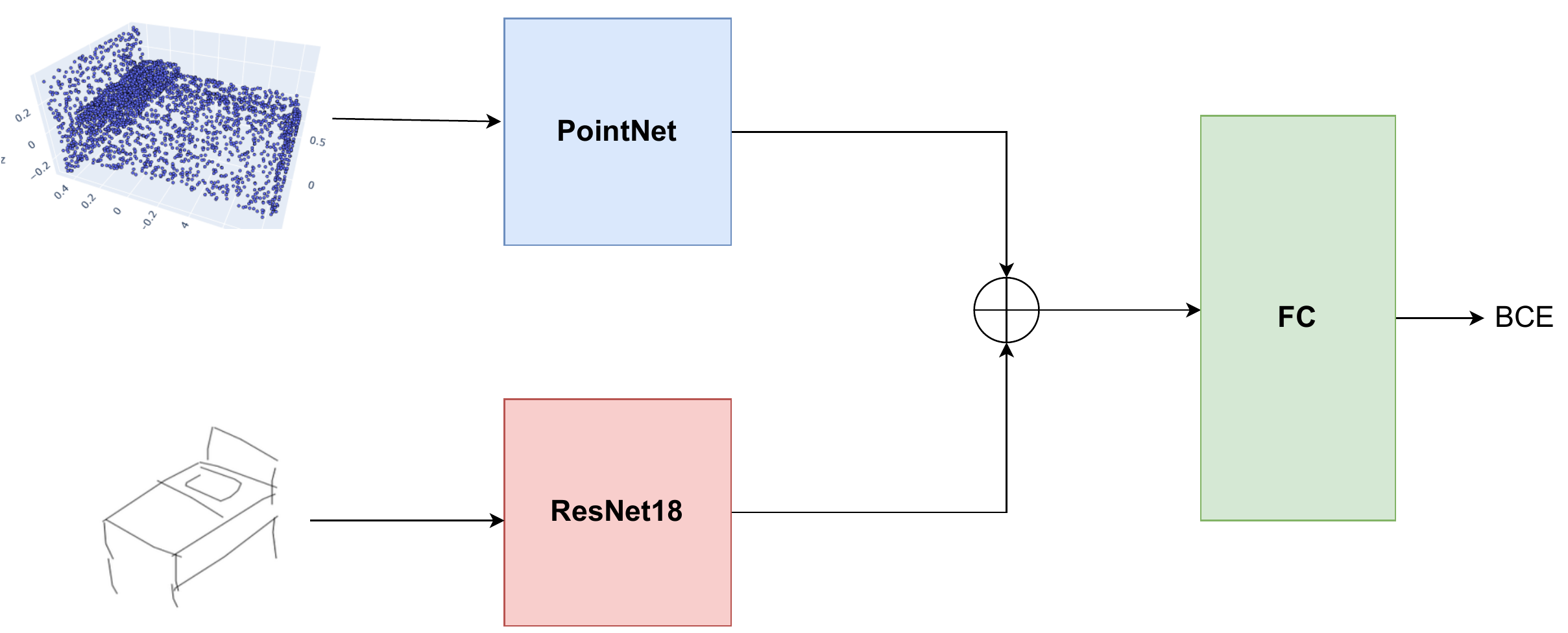}
\caption{Overall framework of Team A's approach.}
\label{fig:pipeline}
\end{figure}

Figure \ref{fig:pipeline} shows the pipeline of our approach. Overall, our approach includes three steps: (1) feature extraction, (2) concatenation, and (3) matching.

\textbf{Sketch Feature Extraction}.
We employ ResNet18 \cite{resnet} to extract deep features for sketches. In practice, we fine-tune several available models pre-trained on ImageNet. The SGD optimizer is adopted with a decay learning rate by a factor of 0.1 every 7 epochs, and the momentum is 0.7. Then, we randomly choose 80\% of the original training set for training and the remaining for validation. We choose the model with the best validation accuracy during the training process.

\textbf{Point Cloud Feature Extraction}.
We use PointNet\cite{pointnet} as the feature extractor for point cloud data. Similar to the training process w.r.t. sketches, we split 80\% of the original training set for training and the rest is used for validation.
Specifically, we use \texttt{sparse\_categorical\_crossentropy} as the loss and the Adam as the optimizer. The learning rate is manually justified after a number of epochs. We choose the model with the best validation accuracy as the one for inference.

As mentioned above, ResNet18 \cite{resnet} and PointNet \cite{pointnet} are employed to extract features of 512 dimensions from sketches and point clouds, respectively. Then, we concatenate the feature vectors from both modalities to create positive and negative samples for training a matching model in the next phase. Two vectors belonging to the same classes in the sketch and the point cloud sets are combined to produce a positive sample, and vectors from different classes are used to generate negative samples. 

We assume that there exists a match between a point cloud and a sketch if their final score is greater than a threshold, which is set as 0.5 in our experiments. Hence, we regard this problem as a binary classification with two classes matching (1) or not matching (0). We train a binary classifier to distinguish positive and negative inputs. The input of the classifier is a concatenated 1024-d vector. The hidden layers are fully connected (FC). The output of this model is a score that represents how the two inputs match with each other.

\begin{figure}[b]
    \centering
    \includegraphics[width=\linewidth]{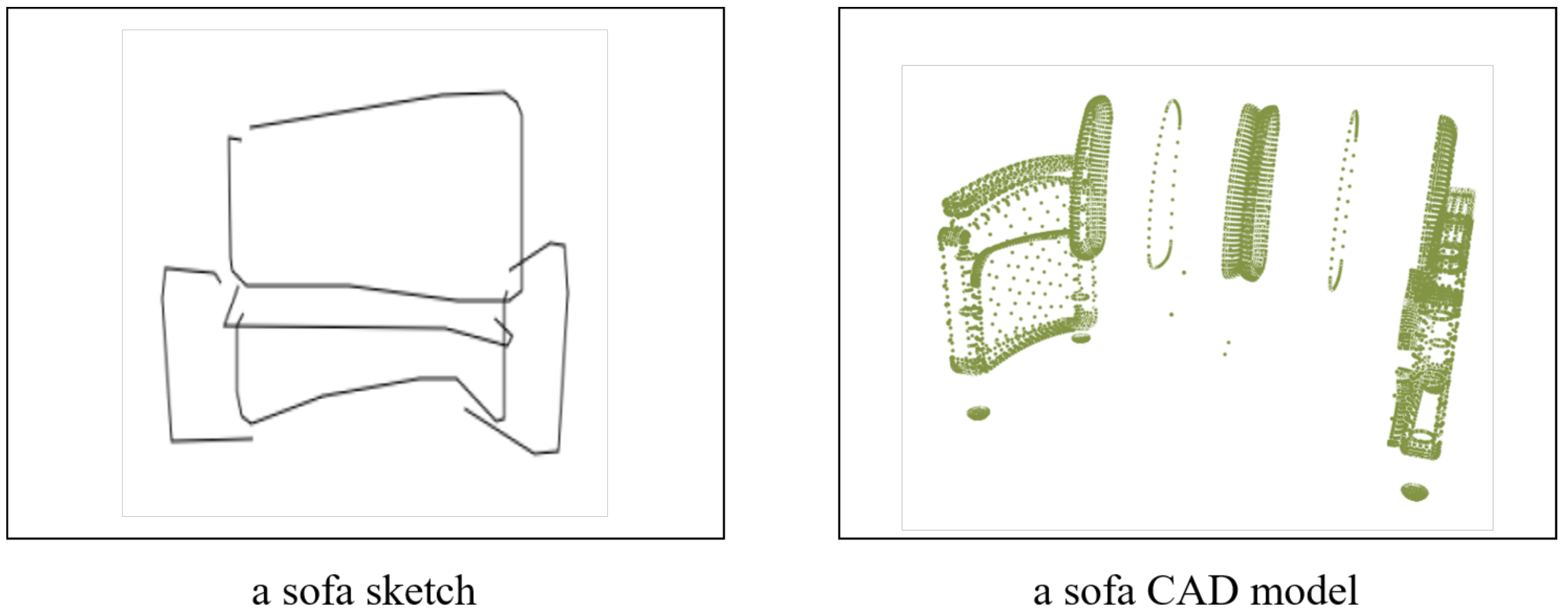}
    \caption{A sketch and a point cloud from the training set within the same class (sofa). The uneven point density makes it hard to deduce its shape from the points. }
    \label{fig:example}
\end{figure}

\subsection{Team B: Common Space Embedding for Sketch-based 3D Shape Retrieval}
\label{sec:hcmus2}

To retrieve 3D models based on 2D sketches, we first note that a sketch generally does not provide an accurate illustration of the models, but describes the concept presented to the drawer \cite{quickdraw}. This gap can be observed in Figure \ref{fig:example}. Therefore, it would be helpful if we can identify the concept described by every sketch and 3D model, and retrieve models with a similar concept to the sketch.

\begin{figure}[t]
    \centering
    \includegraphics[width=\linewidth]{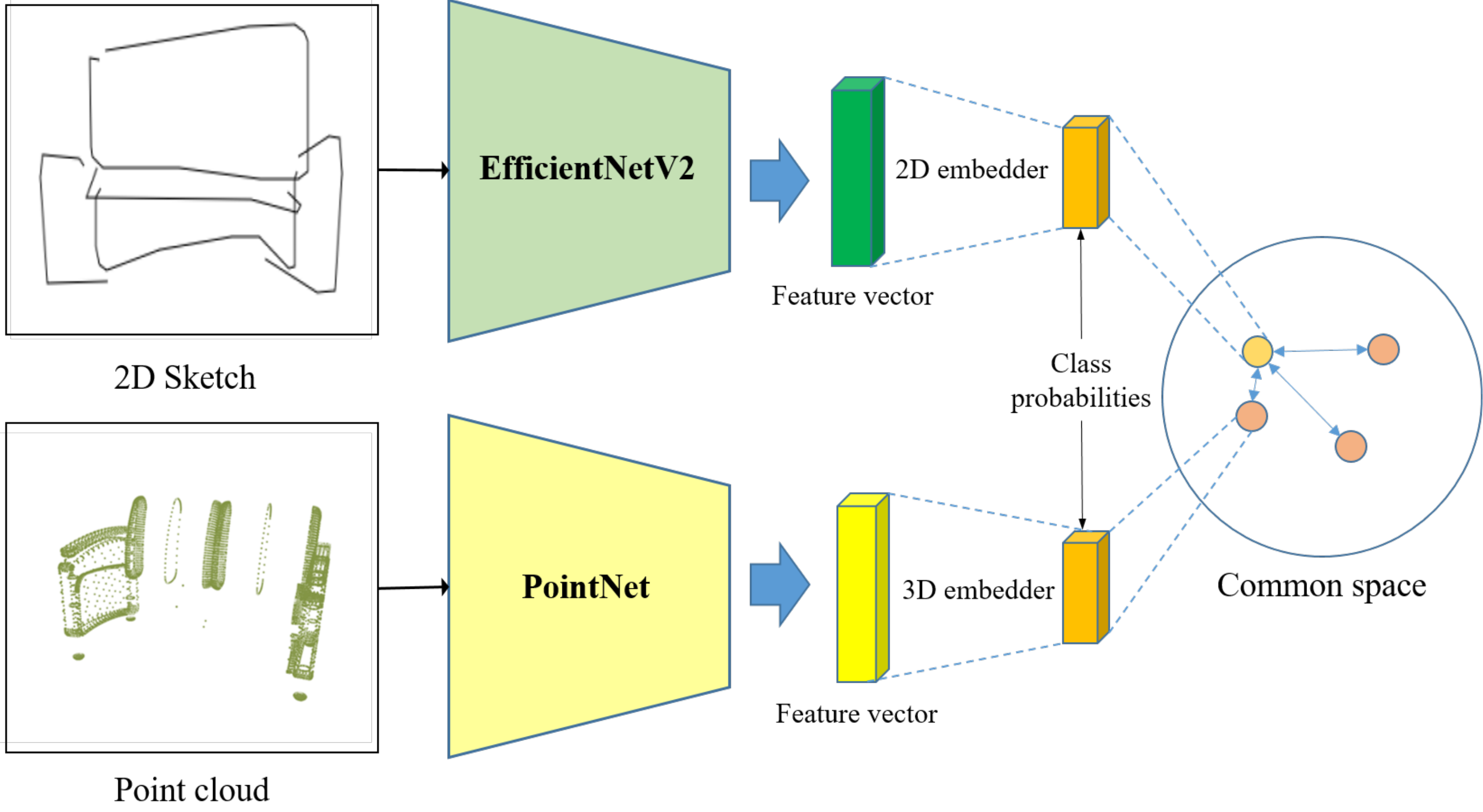}
    \caption{Overall framework of Team B's approach.}
    \label{fig:overview}
\end{figure}

The overview of our method can be seen in Figure \ref{fig:overview}. First, we utilize two feature extractors, one for sketches and the other for 3D models, to obtain features that have $P$ and $Q$ dimensions, respectively. Note that $P$ and $Q$ can be different depending on the models used, creating a gap between the features. To address this, we use two embedding layers to project them to the same $D$-dimensional space. In this space, we can directly compare the representations using cosine similarity. This allows simple and scalable retrieval by methods such as $k$-nearest neighbors.

Note that there are many possible implementations of the aforementioned architecture; for example, we might train the joint embeddings using some form of metric learning. Based on the fact that each sketch/model only describes a single concept, we use the classification probability distribution (\ie, one generated from a softmax function) as the common space. This approach is simple as we can train two classifiers separately, and use the probabilities directly. In the following, we will describe our method and the training process in detail.

\subsubsection{Motivation}
There are a few reasons we do not opt for multi-view approaches. First, the 3D models can be hard to recognize from viewing in 2D. This is because they lack the information of planes and edges, making the process of inferring the shape, and therefore their class, especially challenging even for humans. Trying to reconstruct the meshes is also not trivial since the object can be hollow at arbitrary locations, and the point meshes can be varyingly dense. For example, a table can be represented by as few as 8 points. For those reasons, we choose to directly model the relations between points to find out the semantics of the model. The PointNet model fulfills this requirement; hence we choose it.

\subsubsection{Sketch Model}

We employ the recent EfficientNetV2 \cite{EfficientNetV2} architecture for image feature extraction, using the classification task as the learning objective. This family of models has many different scales and fast convergence speed when training, which are desirable. For each class, we leave out 100 random samples to create the validation sets. We fine-tune available models pretrained on ImageNet using the remaining training sets. We choose small models because they are pretrained with image sizes close to the size of images in the training set ($256 \times 256$). 

We use the RMSprop optimizer with a piecewise constant decay learning rate scheduler that starts at $10^{-3}$ and decays to $10^{-6}$ after 14 epochs. For each task, we pick the model with the best validation accuracy. We note that for these tasks, class imbalance does not cause great troubles, so we do not perform any balancing techniques.

\subsubsection{Point Cloud Model}
\noindent\textbf{Method}.
We use the PointNet \cite{pointnet} model to perform the classification of 3D shapes for 2 tasks. In Task 1, we also apply the voting concept to the model (PointNet $k$-fold) to provide more accurate classification results. The final results will be synthesized from $k$ ($k = 5$ for this task) PointNet models trained separately on different training and test sets.

\noindent\textbf{Training}.
We split the STC dataset into 5 groups, where the samples of each class are divided equally into each group. For each group, we take it as a validation set and take the remaining groups as a training set, based on which we then train a PointNet model. Therefore, there are 5 different PointNet models trained on the STC dataset. On the STW dataset, we take 80\% of the samples as the training set and the rest as the validation set.

\begin{figure*}[htbp]
\centering
\includegraphics[width=0.8\textwidth]{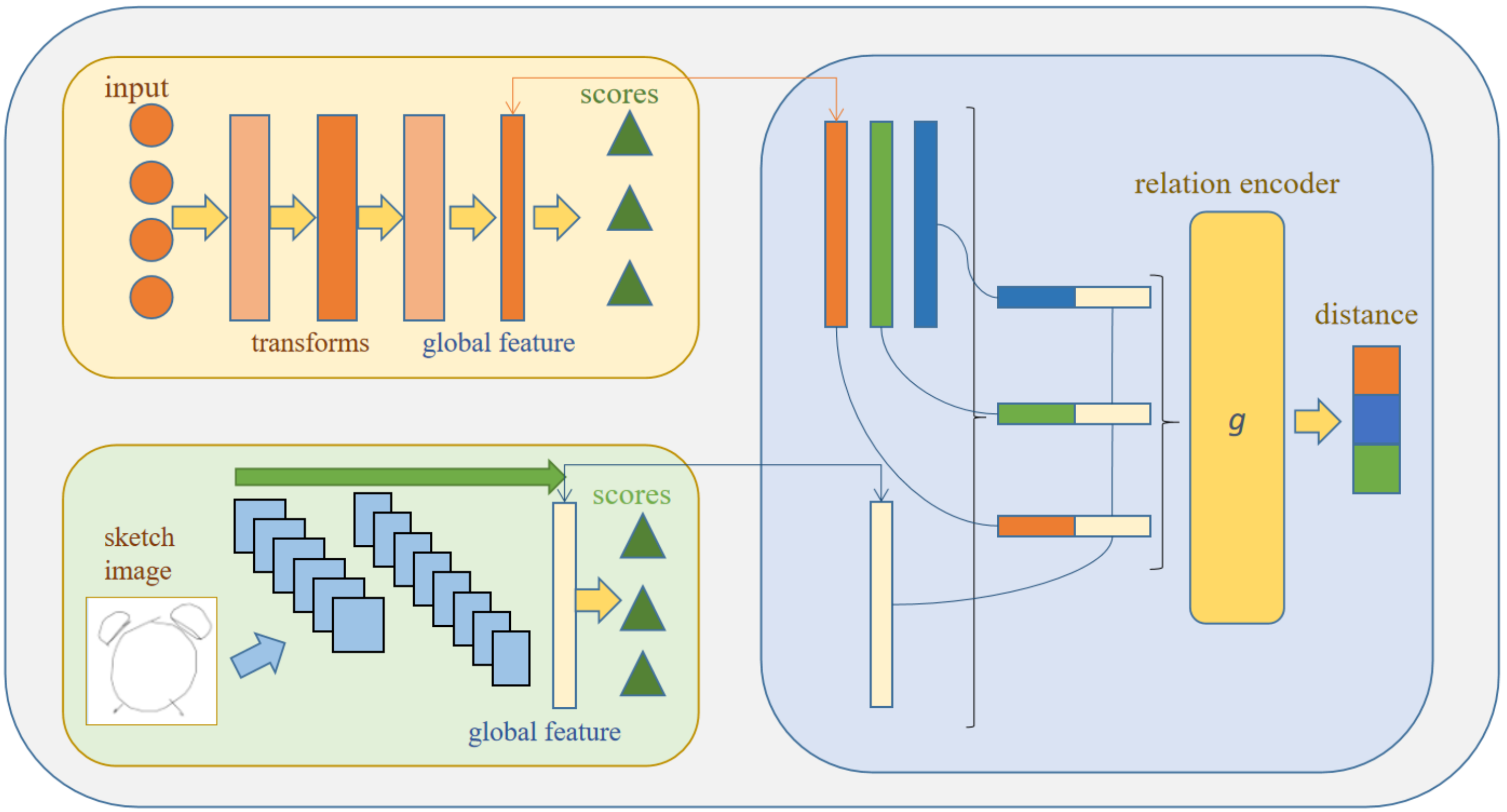}
\caption{Overall framework of Team C's proposed DDRN method. The upper left part is the 2D sketch representation module and the lower left part is the 3D shape branch. The relation evaluation module (right part) takes features from two different modalities as inputs for model construction, and it models the cross-modality co-relation for precise retrieval. The whole network adopts pre-training and fine-tuning paradigm with an optimization in an end-to-end manner.}
\label{fig:ddrn}
\end{figure*}

\noindent\textbf{Implementation Details}. The PointNet models above are trained with the same settings. More details are as follows:
\begin{itemize}
    \item The input of the networks is a point cloud of 1024 points. If a point cloud has less than 1024 points, we clone the points up to sufficient numbers. If a point cloud has more than 1024 points, we randomly select 1024 of those points as input.
    \item We apply the Adam optimizer with a learning rate \\ $\alpha=0.0025$, $\beta_{1}=0.9$, $\beta_{2}=0.999$.
    \item We train each model for 20 epochs with a batch size of 32; finally, we select the weights that give the best accuracy on the corresponding validation set for the inference stage.
\end{itemize}

\subsubsection{Common Space Embedding and Distance-based Retrieval}

As mentioned above, we use the classification probabilities as the common embedding space, \ie, $D = 44$ for Task 1 and $D = 10$ for Task 2. The feature dimensions are $P = 1536$ and $Q = 1024$ for all cases. The embedding layers are fully-connected layers with $D$-dimensional outputs, followed by a softmax activation function.

For each query and each 3D model, we extract their embeddings using our trained models. Then we compute cosine similarity to obtain the distance matrix. Models are ranked by increasing cosine similarity to the sketch query.

We experimented with keeping only the top $k$ ($k = 2, 3, 5$) confidence scores for each sample and zeroing-out the rest, but this resulted in lower metrics on the validation set. This suggests that the full representation is meaningful, even in the low confidence range.

\subsubsection{Runs Description}
\noindent\textbf{Task 1}: We have two runs for this task.
\begin{itemize}
  \item Run 1: Using the best PointNet model to infer the confidence vector (corresponding to 44 classes).
  \item Run 2: Using the PointNet k-fold, and the confidence vector will be synthesized from the average of the results of 44 separately trained models.
\end{itemize}
For both runs, we use a fine-tuned EfficientNetV2-B2 for sketch class probabilities.

\noindent\textbf{Task 2}: We propose the following runs for the second task.
\begin{itemize}
  \item Run 1, 2: Reusing the two models (\ie, PointNet and PointNet k-fold) already running on the STC test set for these runs. Because we want to check if our models trained on the training set of CAD objects can be good enough to recognize 3D objects in the STW test set.  
  \item Run 3: Using the PointMLP \cite{pointMLP} model pre-trained on ScanObjectNN.
  \item Run 4: Using the PointNet model trained on the STW training set.
\end{itemize}
For this task, we fine-tune an EfficientNetV2-B3 with the same procedure as the first task.

\subsection{Team C: Dual-stream Deep Relation Network for Sketch-based 3D Shape Retrieval}
\label{sec:hit}
We propose a dual-stream deep relation network (DDRN) to solve the problem of sketch-based 3D shape retrieval in the wild. To get informative and discriminative representations from two domains, two sub-modules with deep neural networks are employed to extract most representative features of sketches and 3D shapes, respectively.
A cross-modality relation evaluation metric is introduced for measuring the correlation scores among different categories and further guiding the feature extractors. Instead of trivial distance metric, \eg, cosine distance and Euclidean distance, the proposed deep neural relation measurement gives more flexibility and is also helpful to enhance the discrimination of the learning model for the cross-modal retrieval task.
The entire DDRN learning model is trained in a two-step manner, \ie, pre-training and fine-tuning. The overall architecture of this method is shown in Figure~\ref{fig:ddrn}.

\subsubsection{2D Sketch Branch}

Sketches are very different from regular photos. To understand the semantics behind these hand drawing, we adopt the widely-used ResNet-50 \cite{resnet} for deep sketch feature representation learning. The network parameters are initialized by using the pretrained model on ImageNet. To acquire a discriminative representation of the sketches, the pre-training loss of this sub-network is the multi-class cross entropy loss:
\begin{equation}\label{eq1}
\mathcal{L}_{CE}=-\frac{1}{N_{q}} \sum_{i=1}^{N_{q}}\sum_{c=1}^{M_{q}}y_{ic}log(p_{ic}),
\end{equation}
where $N_{q}$ is the total number of sketch samples, $M_{q}$ is the number of classes, $y_{ic}$ is the $c$-th class label of $i$-th sample's label $y_i$, and $p_{ic}$ is the corresponding probability acquired from the network output.
 
\subsubsection{3D Shape Branch}
The network architecture for the 3D shape processing is based on PointNet \cite{pointnet}. Instead of being transformed to regular structure, \ie, 3D voxel grids or 2D aspect projection, the point cloud is fed into the network though its irregularity. The collections of point coordinates go directly through several feature extractors, local and global information aggregation layers, and collaborative alignment networks, which are used for extracting their useful components. Other argumentation strategies such as point permutation invariant are also taken into consideration. We pre-train this module on a classification task for better ability to extract 3D shape features and then fine-tune it in the overall unified training framework. The pre-training loss of this 3D shape network is very similar to the sketch network:
\begin{equation}\label{eq2}
\mathcal{L}_{CE}=-\frac{1}{N_{z}} \sum_{i=1}^{N_{z}}\sum_{c=1}^{M_{z}}y_{ic}log(p_{ic}),
\end{equation}
where $N_{z}$ is the total number of shape samples, and $M_{z}$ is the number of classes.

\subsubsection{Relation Network}
Inspired by the well-known Relation Network \cite{relation_net}, we propose a relation distance encoder to construct the correlated relationship between sketch image and retrieved 3D shape features.

Firstly, after the aforementioned sketch and 3D shape modal-encoders, we obtain the global feature $\bm q$ of the queried sketch image and the global feature ${\bm z_i}$ of the retrieved 3D shapes by utilizing the output of the feature layer before classification. In order to obtain the distance matrix between each query point $\bm q$ and the retrieved 3D shapes $\bm z_i$, this model is also required to be capable of measuring the distance $d_i$ between the query point $\bm q$ and each 3D shape sample $\bm z_i$. Herein, we simply concatenate the query point $\bm q$ with each 3D shape samples to form several query-sample pairs $(\bm q, \bm z_i)$. For each query sample pair, following \cite{relation_net}, we put it into a two-layer multi-layer perceptron (MLP). The activation function of the first layer is the linear rectification function(ReLU) and the sigmoid function is attached to the linear out put as the activation function of the second layer. In this way, the given algorithm can obtain the similarity score between the query point and the simulated retrieval samples $s_i \in (0,1)$:
\begin{equation}\label{eq3}
s_i = \sigma(\bm W_2(ReLU(\bm W_1(\bm q \| \bm z_i)+\bm b_1))+\bm b_2),
\end{equation}
where $\sigma$ is the sigmoid function. $\bm W_1$, $\bm W_2$, $\bm b_1$, and $\bm b_2$ are the network parameters, and $\|$ is the concatenation operator. The objective of the relation distance network is modeled as:
\begin{equation}\label{eq4}
\min_{\bm W_1, \bm W_2, \bm b_1, \bm b_2} \sum_{i=1}^m \|\bm s_i - \bm 1(c(\bm q)=c(\bm z_i))\|^2,
\end{equation}
where $m$ is the number of sampled classes per batch, $c(\cdot)$ returns the class label of a datapoint, and $\bm 1(\cdot)$ has value $1$ when the condition is true, or has value $0$, otherwise. This objective gives features of the same category a higher similarity score, while lower scores are achieved when features are from different classes.
In this work, we use ${d_i = 1 - s_i}$ as the distance score between the retrieved samples and the query point, so that the query sample pairs with large (small) similarities have close (far) distances. The relation distance network can explore the correlation degree of features from different modalities.

\subsubsection{Optimization}
The overall optimization process of our method is simply summarized in these steps for the two tasks:
\begin{itemize}
  \item Step 1: Pre-train the ResNet-50 on a sketch dataset using Eq. (\ref{eq1}).
  \item Step 2: Pre-train the PointNet on a corresponding 3D shape dataset using Eq. (\ref{eq2}).
  \item Step 3: Acquire the discriminative representations of the two modals with the trained two sub-networks.
  \item Step 4: Train the relation distance network with Eq. (\ref{eq4}).
\end{itemize}

\subsection{Team D: Suggestive Contours with Domain-aware Squeeze-and-Excitation for Sketch-based 3D Shape Retrieval}
\label{sec:cczu}
We propose a new sketch-based 3D shape retrieval model to bridge the gap between sketches and 3D point clouds. This network is divided into the suggestive contour of generating line drawing from 3D models, and the DASE network \cite{2018Domain} using SE module and multiplicative Euclidean margin softmax loss.

\begin{figure}[!t]
\centering
\includegraphics[width=0.95\linewidth]{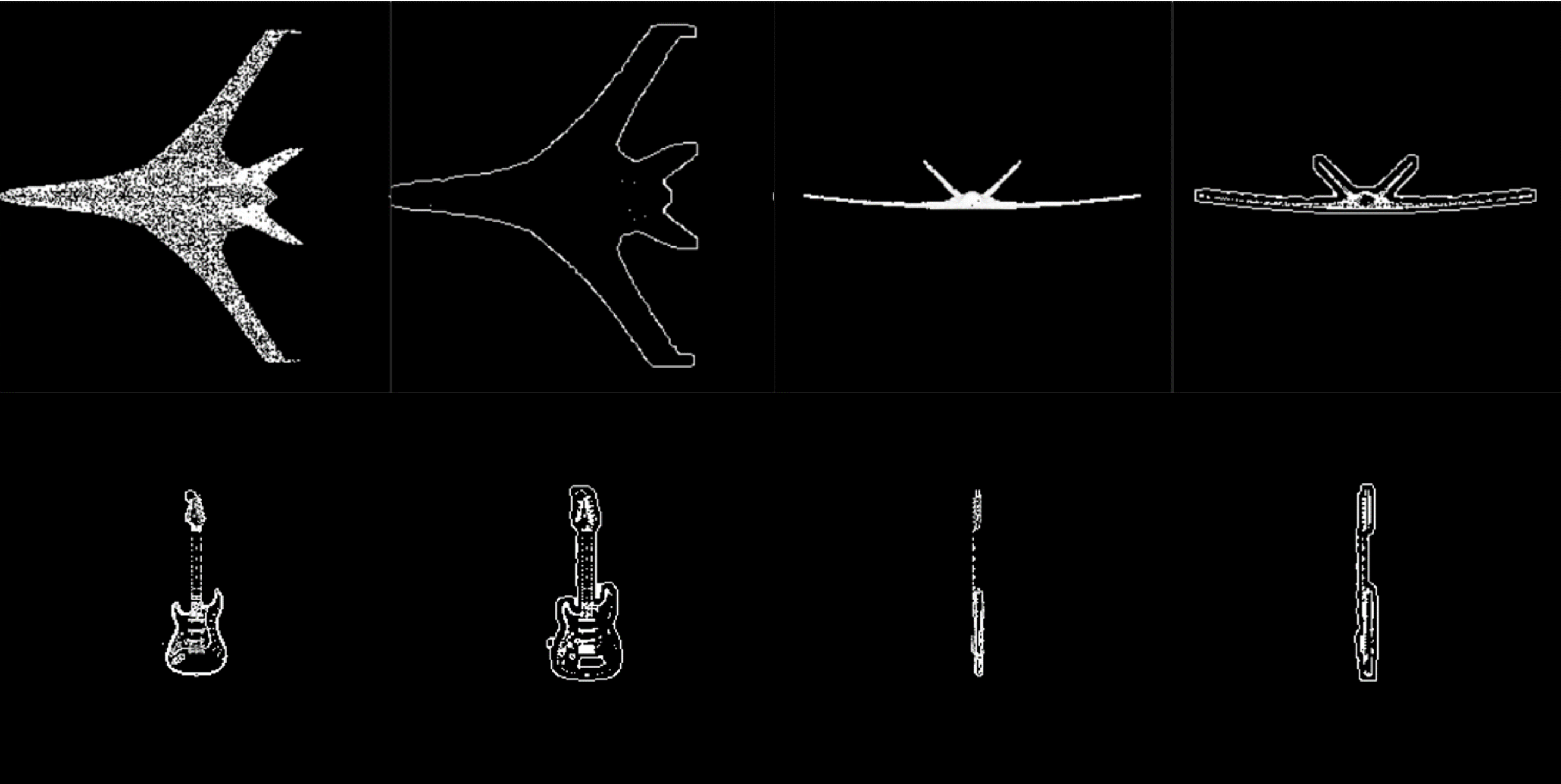}
\caption{Different views of the 3D point clouds. From left to right: frontal projection images of point cloud, suggestive contours images of frontal projection, side projection images of point cloud, suggestive contours images of side projection.}
\label{fig:view}
\end{figure}

Learning common feature representations for sketch and 3D point cloud domains is non-trivial. As shown in Figure \ref{fig:view}, sketches are iconic and abstract with various deformation levels; meanwhile, photos are realistic images with shape information. It is distinct what edges are in an image by traditional image process algorithms such as Canny, but it is difficult to 3D point clouds. To draw an accurate line drawing of a 3D point cloud, the algorithm must understand the 3D shape, and select lines which can represent the shape of the 3D point cloud. We thus adopt a method \cite{DeCarlo:2003:SCF} for generating line drawing from point clouds.

\begin{figure}[!b]
\centering
\includegraphics[width=0.5\linewidth]{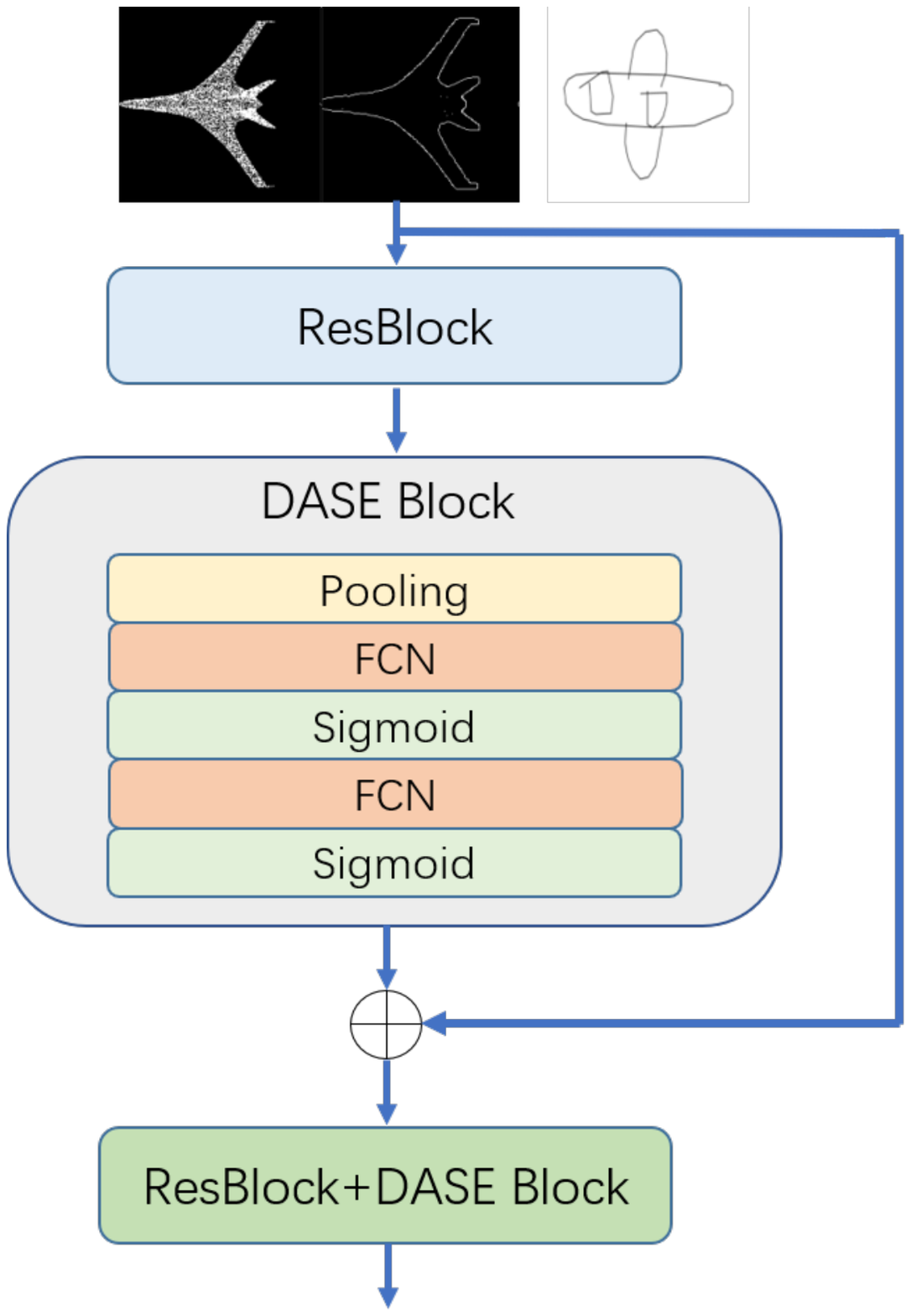}
\caption{The structure of the DASE module.}
\label{fig:dase}
\end{figure}
Our model learns a single network to analyze the projection of the 3D point clouds and sketches jointly rather than using independent sub-networks to process the projection images and sketches separately. The SiameseNet-like network is efficient in learning the embedding space across different domains. Liu \emph{et al.} \cite{2017Deep} proposed a semi-heterogeneous network, including a shared Siamese sub-network and an independent sub-network for each domain.

\begin{table*}[!t]
\centering
\caption{Performance (in \%) comparison on the STC benchmark for sketch-based 3D CAD model retrieval (Task 1).}
\begin{tabular}{lccccccc}
\hline
Team & \#Run & NN & FT & ST & E & DCG & mAP\\
\hline\hline
\multirow{2}{*}{Team B~~~}  &  ~~~1~~~ &  ~~~92.25~~~ &  ~~~85.89~~~ &  ~~~92.06~~~ & ~~~48.50~~~ & ~~~95.11~~~ & ~~~89.28~~~ \\ 
\cline{2-8}
& 2 & 92.23 & 86.96 & 92.77 & 49.04 & 95.40 & 90.18 \\ 
 \hline
Team C & 1 & 1.08 & 1.54 & 3.10 & 0.11 & 36.29 & 2.05 \\ 
 \hline
 \multirow{3}{*}{Team D}  & 1 & 2.35 & 1.94 & 3.92 & 0.36 & 38.16 & 2.23 \\
 \cline{2-8}
 & 2 & 2.63 & 1.80 & 3.62 & 0.16 & 37.57 & 1.99 \\
 \cline{2-8}
 & 3 & 2.10 & 1.95 & 3.90 & 0.35 & 37.84 & 2.03 \\
 \hline
Baseline-MV & 1 & 2.91 & 1.82 & 3.56 & 0.34 & 37.65 & 2.00 \\
 \hline
Baseline-PC & 1 & 1.65 & 1.85 & 4.04 & 0.33 & 37.90 & 2.01 \\
 \hline
\end{tabular}
\label{tab:task1}
\end{table*}
\begin{table*}[!t]
\centering
\caption{Performance (in \%) comparison on the STW benchmark for sketch-based realistic scanned model retrieval (Task 2). * The first 3 runs of Team B are just for reference as they use auxiliary data for training.}
\begin{tabular}{lccccccc}
\hline
Team & \#Run & NN & FT & ST & E & DCG & mAP\\
\hline\hline
Team A & 1 & 39.73 & 44.71 & 63.10 & 14.47 & 77.17 & 46.67 \\ 
 \hline
\multirow{4}{*}{Team B~~~}  &  ~~~1*~~~ &  ~~~1.19~~~ &  ~~~12.55~~~ &  ~~~22.63~~~ & ~~~4.78~~~ & ~~~60.57~~~ & ~~~13.43~~~ \\
\cline{2-8}
& 2* & 1.05 & 7.97 & 18.69 & 1.84 & 57.37 & 10.26 \\
\cline{2-8}
& 3* & 0.34 & 7.67 & 16.94 & 1.09 & 56.5 & 9.99 \\
\cline{2-8}
& 4 & 71.16 & 61.29 & 71.81 & 25.18 & 86.18 & 67.31 \\
 \hline
 Team C & 1 & 10.93 & 11.13 & 20.58 & 3.86 & 60.18 & 15.15 \\
 \hline
 \multirow{3}{*}{Team D}  & 1 & 10.23 & 9.85 & 19.52 & 3.08 & 58.75 & 10.09 \\
 \cline{2-8}
 & 2 & 9.18 & 9.67 & 19.51 & 3.18 & 58.75 & 10.05 \\
 \cline{2-8}
 & 3 & 9.94 & 9.73 & 19.45 & 3.03 & 58.73 & 10.09 \\
 \hline
Baseline-MV & 1 & 9.32 & 13.10 & 26.89 & 6.08 & 61.07 & 15.52\\
 \hline
Baseline-PC & 1 & 9.32 & 13.74 & 28.38 & 7.59 & 62.69 & 18.81 \\
 \hline
\end{tabular}
\label{tab:task2}
\end{table*}

We add the Domain-Aware Squeeze-and-Excitation (DASE) module after each residual block, for learning an effective feature representation to provide an explicit mechanism for re-weighting the importance of channels. As shown in Figure \ref{fig:dase}, the DASE module has an encoder-decoder structure and the sigmoid activation. And we append one binary value code indicates whether comes from the sketcher domain or the suggestive contours images of projection domain, to the low dimensional embedding. After the sigmoid activation, we can get the feature attention vector. Therefore, this conditional structure can help capture different features of the input image.

The SE module proposed by \cite{2017Squeeze}, using an encoder network to squeeze the convolution features into a low dimensional embedding, and the characteristic responses of channels are adaptively recalibrated by explicitly simulating the interdependence between different channels. So the DASE module learned between two domains can explore the shared semantics in a common feature space.

The loss function plays an important role in optimizing the networks over the recognition tasks successfully, particularly in our cross-domain scenario. In order to embed photos and sketches into the shared space, we use a Multiplicative Euclidean Margin Softmax (MEMS) loss, which utilizes the strategy of maximizing intra-class distance and minimizing inter-class distance.

Specifically, the maximum intra-class distance can be defined by $\max_{x,x' \in D_y} x'\in (\cup_{y'\neq y}D_{y'})$and the minimum inter-class distance would be min $x\in D_y, x'\in (\cup_{y'\neq y}D_{y'})d(x, x')$, where $d(\ \cdot\ )$ can be any kind of differentiable distance metric if the category $y$ is given. Therefore, we optimize our framework by forcing the maximum intra-class distance to be less than the minimum inter-class distance to develop our objective.
\begin{equation}
\label{eq:8}
\max _{\boldsymbol{x}, \boldsymbol{x}^{\prime} \in \mathcal{D}_{y}} \mathrm{~d}\left(x, x^{\prime}\right) \leq \min _{\boldsymbol{x} \in \mathcal{D}_{y}, \boldsymbol{x}^{\prime} \in\left(\cup_{y^{\prime} \neq y} \mathcal{D}_{y^{\prime}}\right)} \mathrm{d}\left(x, x^{\prime}\right)
\end{equation}
We compute the prototypes $\{c_y\} \subset \chi $of each class, and characterize the distribution of instances in feature space. Instances will be closer to their corresponding prototypes than others in feature space if the MEMS loss is optimized well.
\begin{equation}
\label{eq:9}
\forall\left(x, y_{x}\right) \in \mathcal{D},\  m \cdot \mathrm{d}\left(x, \boldsymbol{c}_{y_{x}}\right) \leq \mathrm{d}\left(x, \boldsymbol{c}_{y}\right)
\end{equation}
where $m \geq 1$ is referred to as the margin constant.

\begin{figure*}[!t]
\begin{center}
   \includegraphics[width=\linewidth]{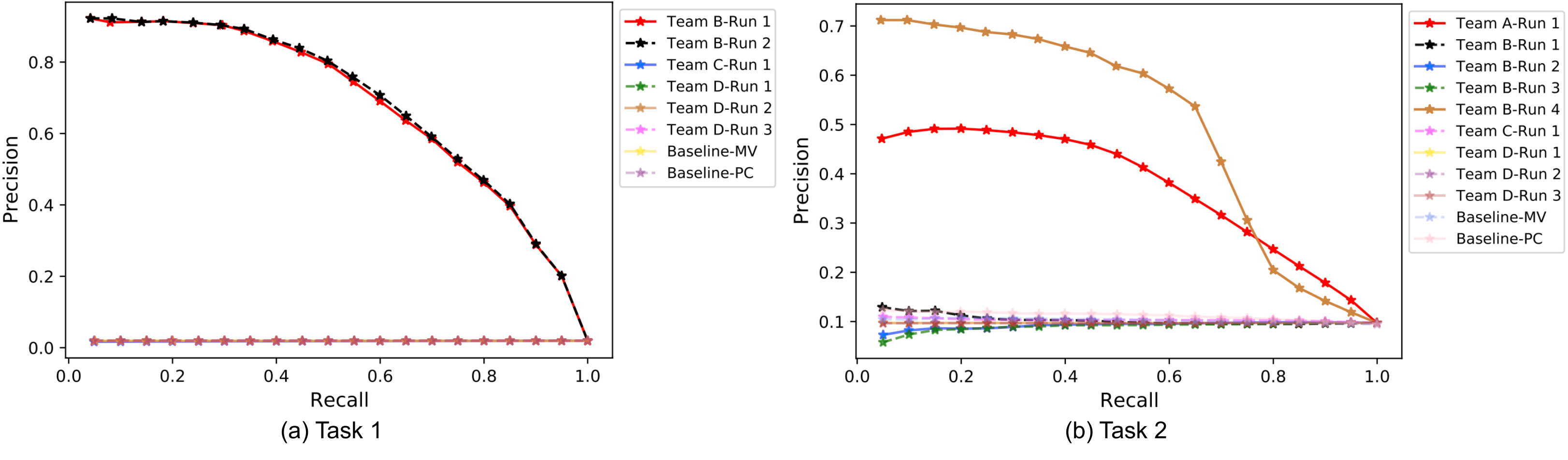}
\end{center}
\vspace{-5mm}
   \caption{Precision-recall curves for all the runs of the 4 teams in terms of (a) Task 1 and (b) Task 2.}
\label{fig:pr_curve_all}
\end{figure*}
\begin{figure*}[!t]
\begin{center}
   \includegraphics[width=\linewidth]{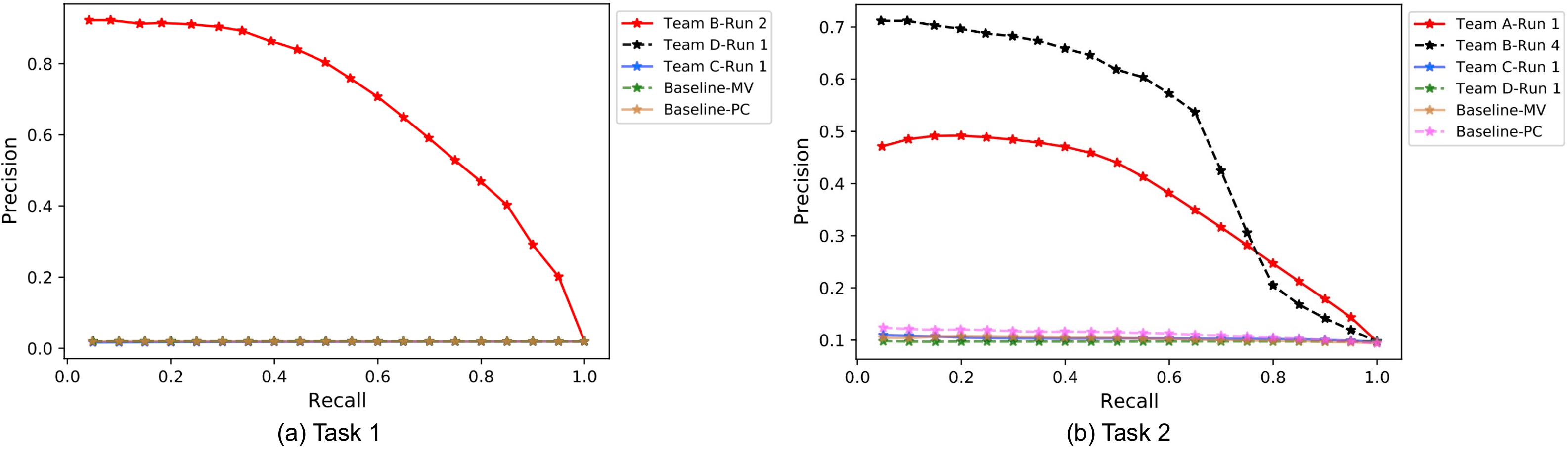}
\end{center}
\vspace{-5mm}
   \caption{Precision-recall curves for the best runs of the 4 teams in terms of (a) Task 1 and (b) Task 2.}
\label{fig:pr_curve_best}
\end{figure*}

We denote $R_{y,y'}$ as a region in the feature space for convenience, and $R_y$ as a region such that:
\begin{equation}
x \in \mathcal{R}_{y, y^{\prime}} \Leftrightarrow m \cdot \mathrm{d}\left(x, c_{y}\right) \leq \mathrm{d}\left(x, c_{y^{\prime}}\right)
\end{equation}
\begin{equation}
x \in \mathcal{R}_{y} \Leftrightarrow  m \cdot \mathrm{d}\left(x, \boldsymbol{c}_{y}\right) \leq \mathrm{d}\left(x, \boldsymbol{c}_{y^{\prime}}\right)
\end{equation}
which considers different data distributions of each class. $\mathcal{R}_{y}=\bigcap_{y^{\prime} \neq y} \mathcal{R}_{y, y^{\prime}}$ is easy to prove. If Eq. (\ref{eq:9}) holds, we can derive a sufficient condition for Eq. (\ref{eq:8}):
\begin{equation}
\forall y \in \mathcal{Y}, \max _{x, \boldsymbol{x}^{\prime} \in \mathcal{R}_{y}} \mathrm{~d}\left(x, x^{\prime}\right) \leq \min _{x \in \mathcal{R}_{y}, x^{\prime} \in\left(\cup_{y^{\prime} \neq y} \mathcal{R}_{y^{\prime}}\right)} \mathrm{d}\left(x, x^{\prime}\right) .
\end{equation}
To incorporate this spirit and optimize Eq. (\ref{eq:9}) by the softmax loss, this gives us a novel loss function, \ie, the MEMS loss:
\begin{equation}
\mathcal{L}_{MEMS}=\frac{1}{N} \sum_{i=1}^{N}-\log \frac{e^{-m^{2}\left\|x_{i}-c_{y_{i}}\right\|_{2}^{2}}}{e^{-m^{2}}\left\|x_{i}-c_{y_{i}}\right\|_{2}^{2}+\sum_{j \neq y_{i}} e^{-\left\|x_{i}-c_{j}\right\|_{2}^{2}}}
\end{equation}
where $x_i$ indicates the feature extracted by the last layer of feature representation network. The center of $j$-th category is $c_j$. Rather than directly using the average feature center, we take the center $c_j$ as the parameters and update $c_j$ dynamically. We employ the negative squared Euclidean distance to measure the confidence of $x_i$ being $\left(-\left\|x_{i}-c_{j}\right\|_{2}^{2}\right)$. Particularly, in binary classification, $x_{i}$ is labelled as class 1 if $m\left\|x-c_{1}\right\|_{2}<\left\|x-c_{2}\right\|_{2}$, and otherwise, as class 2.

\section{Results and Discussions}
\label{result}
In this section, we first show the comparative results of the two baseline methods on the STC and STW benchmarks, and then demonstrate the evaluation results for each task, respectively, based on the distance matrices submitted by the four teams. The retrieval performance is measured by the seven metrics (\ie, NN, FT, ST, E, DCG, mAP, PR) as introduced in Section \ref{sec:eval}.

\subsection{Baseline Results}
\begin{figure}[!t]
\begin{center}
   \includegraphics[width=0.8\linewidth]{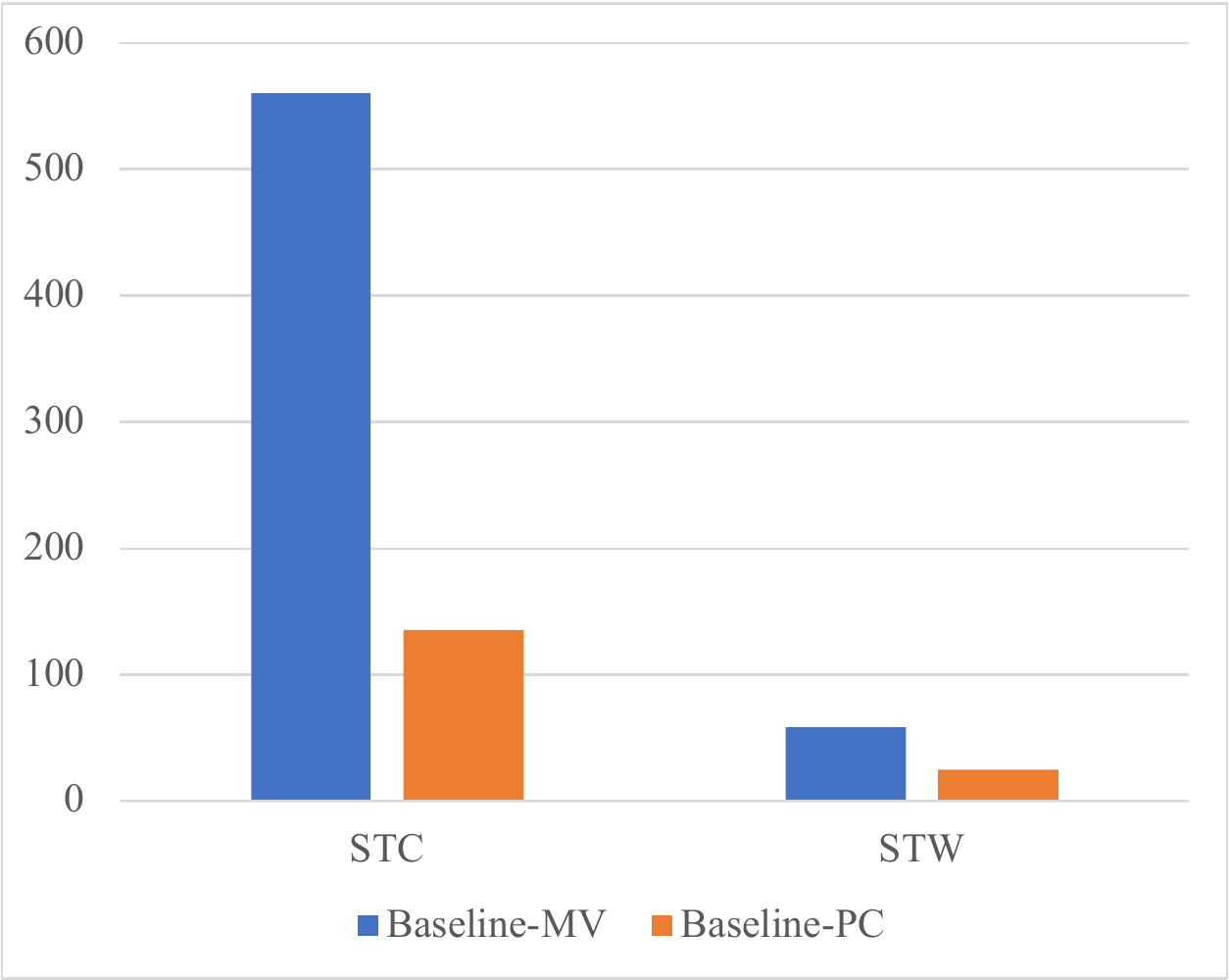}
\end{center}
\vspace{-5mm}
   \caption{Runtime (in seconds) comparison between the two baseline methods on the STC and STW benchmarks.}
\label{fig:runtime}
\end{figure}

We show the evaluation results of the two baseline methods for the two tasks in Tables \ref{tab:task1} and \ref{tab:task2}, respectively. From the tables, we can see that both baselines achieve comparable but poor performance on the STC benchmark. This is probably due to the large scale of this dataset which consists of $\sim$120k sketches in total. On the STW benchmark, Baseline-PC slightly outperforms Baseline-MV, indicating that point cloud based representations are relatively better than multi-view representations on this realistic dataset. We also show the precision-recall curves in Figures \ref{fig:pr_curve_all} and \ref{fig:pr_curve_best}, where we have the similar observations. Moreover, the retrieval time during the test phase is reported in Fig. \ref{fig:runtime}. We can see that Baseline-PC is much more efficient than Baseline-MV, by avoiding the time-consuming rendering process. For instance, Baseline-PC is around 5 times faster than Baseline-MV on STC. This indicates that directly representing 3D point clouds not only mimics the realistic setting but also is more practical in real applications.

\subsection{Results for Task 1}
Since Task 1 is of much larger scale than Task 2, there are only 3 teams participating in this challenging task. We show the detailed comparative results in Table \ref{tab:task1} and Figures \ref{fig:pr_curve_all}(a) and \ref{fig:pr_curve_best}(a). As can be seen from the table and figures, Team C and Team D fail to obtain satisfactory performance, which is comparable to that of the two baseline methods. Team B performs the best and there exists a large performance gap between the team and the other two. For instance, their 2nd run achieves over 90\% in mAP. This is probably due to that Team B employs a common embedding space of the same dimension with the number of categories of sketches/shapes. In other words, once the classification network is optimized, the common embedding is equivalent to the classification probabilities, and is thus very effective.

\subsection{Results for Task 2}
Table \ref{tab:task2} and Figures \ref{fig:pr_curve_all}(b) and \ref{fig:pr_curve_best}(b) illustrate the performance obtained by 9 runs of 4 teams. Note that the first 3 runs of Team B are just for reference as they use auxiliary data for training. From the table and figures, we can see that, once again, Team B outperforms all the other teams by a large margin. The model trained on the STW benchmark (\ie, Team B-run 4) is superior to those trained using auxiliary data (\ie, Team B-run 1-3), showing the differences between our realistic dataset with others. Furthermore, Team B's performance for Task 2 is lower than Task 1, indicating the challenges of retrieving 3D shapes in the wild. Team C achieves comparable performance with Baseline-MV, but underperforms Baseline-PC. And Team D ranks the last - their performance is lower than both baseline methods. Interestingly, Team C and Team D achieve higher mAPs for Task 2 than the ones for Task 1. This is probably because the scale of the STW dataset is much smaller than that of STC, and more importantly, all the 3D models on STW are involved during both the training and the test processes.

\subsection{Discussions}
From the results above, we can see that Team C and Team D obtain unsatisfactory results for both tasks. In some cases, their results are lower than the point cloud based baseline method (\ie, Baseline-PC). After looking into the data themselves, one cause of this defect may turn out to be the class imbalance, which is especially obvious on the STC benchmark. Team C relies on the constructed positive and negative pairs to train the relation network. Due to the imbalance, the positive pairs are overwhelmed by the negative ones, making the model hard to capture useful similarity information. One suggestion for them to improve the performance may be developing a better sampling method to generate appropriately balanced pairs, or simply changing the weighting factors between positive and negative pairs. Another reason that leads to the unsatisfactory performance could be that point clouds are too abstract and cannot represent the shape as detailed as meshes, which are available on previous SBSR datasets. In such cases, local geometric details may be lost after 2D projections. Therefore, we suggest Team D to further integrate 3D point cloud representations into their framework to make the overall representations more robust and discriminative. In addition to the challenging benchmarks themselves, the unsatisfactory performance may be also related to the design of their methods. In terms of Team C’s method, the core component is the relation encoder that pays specific attention to the relationship between the two modalities. However, it unintentionally omits the discriminability of the individual modality. Although cross-entropy losses are used during pre-training, such classification losses are still needed to ensure the learned features from the corresponding modality to be more discriminative when training the whole DDRN network. As for Team D’s approach, we observe that their framework focuses on learning discriminative feature representations for both modalities through a single network architecture. In this case, there is no interaction/alignment between the two modalities, thus leading to inferior results. In summary, both the challenging benchmarks and the proposed frameworks lead to the unsatisfactory results. We encourage them to focus on the discriminability of the single modality as well as the alignment between the two modalities when designing the SBSR frameworks in the future.

\section{Conclusions}
Encouraged by the success of previous SHREC tracks on sketch-based 3D shape retrieval, this track aims to further foster this important research theme by introducing two realistic and challenging tasks and the corresponding STC and STW benchmarks. Four teams have participated in the two tasks and contributed 15 runs of their proposed methods. Overall, this track provides large-scale benchmarks for evaluating existing sketch-based 3D shape retrieval methods and at the same time for fostering novel approaches and new research directions by mimicking real-world application scenarios.

In future works, from the methodology perspective, learning from imbalanced data deserves further exploration, especially for the first task as the number of CAD shapes per category varies significantly. Besides, how to learn a common/shared latent space that is able to align the representations of the two modalities well is also a possible direction for improving the retrieval performance. Last but not least, for the realistic 3D models, a possible way for performance improvement is to refine the 3D models by noise removal or data augmentation techniques. From the challenge perspective, we plan to expand the size of the STW dataset by adopting other realistic 3D datasets. Another future direction could be exploring the 2D-3D cross-modality retrieval in other application scenarios, such as scene retrieval. Furthermore, in addition to the retrieval task, the challenge can also be extended to other settings, such as 2D sketch-based 3D shape synthesis/generation.

\section*{Acknowledgments}
We sincerely thank the organizers of SHREC 2022 and 3DOR 2022 for their help in the organization of this track. We thank the researchers who collected the sketch and 3D shape benchmarks used in this track. We thank the anonymous reviewers for providing constructive comments, which helped us to improve and clarify this work. We also thank Peng Zheng who helped us to perform the evaluation.

This work was partially supported by the Fundamental Research Funds for the Central Universities (No. NS2022083). This work was also partially supported by the NYUAD Center for Artificial Intelligence and Robotics (CAIR), funded by Tamkeen under the NYUAD Research Institute Award CG010. The work of Shuaihang Yuan was supported by the NYUAD Global Ph.D. Student Fellowship, funded by the Inception Institute of Artificial Intelligence. The team from Changzhou University was supported by the National Natural Science Foundation of China (No. 61976028). The work of the teams from University of Science, VNU-HCM (HCMUS) was funded by Vingroup and supported by Vingroup Innovation Foundation (VINIF) under project code VINIF.2019.DA19.
%%Vancouver style references.
\bibliographystyle{cag-num-names}
\bibliography{refs}

\begin{thebibliography}{26}
\providecommand{\natexlab}[1]{#1}
\providecommand{\url}[1]{\texttt{#1}}
\providecommand{\href}[2]{#2}
\providecommand{\path}[1]{#1}
\providecommand{\eprint}[1]{\href{http://arxiv.org/abs/#1}{\path{#1}}}
\providecommand{\DOIprefix}{doi:}
\providecommand{\ArXivprefix}{arXiv:}
\providecommand{\URLprefix}{URL: }
\providecommand{\Pubmedprefix}{pmid:}
\providecommand{\doi}[1]{\href{http://dx.doi.org/#1}{\path{#1}}}
\providecommand{\Pubmed}[1]{\href{pmid:#1}{\path{#1}}}
\providecommand{\BIBand}{and}
\providecommand{\bibinfo}[2]{#2}
\ifx\xfnm\undefined \def\xfnm[#1]{\unskip,\space#1}\fi
%Type = Article
\bibitem[{Li et~al.(2014{\natexlab{a}})Li, Lu, Godil, Schreck, Bustos, Ferreira
  et~al.}]{li2014comparison}
\bibinfo{author}{Li\xfnm[ B]}, \bibinfo{author}{Lu\xfnm[ Y]},
  \bibinfo{author}{Godil\xfnm[ A]}, \bibinfo{author}{Schreck\xfnm[ T]},
  \bibinfo{author}{Bustos\xfnm[ B]}, \bibinfo{author}{Ferreira\xfnm[ A]},
  et~al.
\newblock \bibinfo{title}{A comparison of methods for sketch-based 3d shape
  retrieval}.
\newblock \bibinfo{journal}{Computer Vision and Image Understanding}
  \bibinfo{year}{2014}{\natexlab{a}};\bibinfo{volume}{119}:\bibinfo{pages}{57--80}.
%Type = Inproceedings
\bibitem[{Chen et~al.(2019)Chen, Qin, Liu, Zhu, Shen, Xie
  et~al.}]{chen2019deep}
\bibinfo{author}{Chen\xfnm[ J]}, \bibinfo{author}{Qin\xfnm[ J]},
  \bibinfo{author}{Liu\xfnm[ L]}, \bibinfo{author}{Zhu\xfnm[ F]},
  \bibinfo{author}{Shen\xfnm[ F]}, \bibinfo{author}{Xie\xfnm[ J]}, et~al.
\newblock \bibinfo{title}{Deep sketch-shape hashing with segmented 3d
  stochastic viewing}.
\newblock In: \bibinfo{booktitle}{Proceedings of the IEEE/CVF Conference on
  Computer Vision and Pattern Recognition}. \bibinfo{year}{2019}, p.
  \bibinfo{pages}{791--800}.
%Type = Inproceedings
\bibitem[{Chen and Fang(2018)}]{chen2018deep}
\bibinfo{author}{Chen\xfnm[ J]}, \bibinfo{author}{Fang\xfnm[ Y]}.
\newblock \bibinfo{title}{Deep cross-modality adaptation via semantics
  preserving adversarial learning for sketch-based 3d shape retrieval}.
\newblock In: \bibinfo{booktitle}{Proceedings of the European Conference on
  Computer Vision (ECCV)}. \bibinfo{year}{2018}, p. \bibinfo{pages}{605--620}.
%Type = Inproceedings
\bibitem[{Li et~al.(2013)Li, Lu, Godil, Schreck, Aono, Johan
  et~al.}]{li2013shrec}
\bibinfo{author}{Li\xfnm[ B]}, \bibinfo{author}{Lu\xfnm[ Y]},
  \bibinfo{author}{Godil\xfnm[ A]}, \bibinfo{author}{Schreck\xfnm[ T]},
  \bibinfo{author}{Aono\xfnm[ M]}, \bibinfo{author}{Johan\xfnm[ H]}, et~al.
\newblock \bibinfo{title}{Shrec’13 track: large scale sketch-based 3d shape
  retrieval}.
\newblock In: \bibinfo{booktitle}{Eurographics Workshop on 3D Object
  Retrieval}. \bibinfo{year}{2013},.
%Type = Inproceedings
\bibitem[{Li et~al.(2014{\natexlab{b}})Li, Lu, Li, Godil, Schreck, Aono
  et~al.}]{li2014shrec}
\bibinfo{author}{Li\xfnm[ B]}, \bibinfo{author}{Lu\xfnm[ Y]},
  \bibinfo{author}{Li\xfnm[ C]}, \bibinfo{author}{Godil\xfnm[ A]},
  \bibinfo{author}{Schreck\xfnm[ T]}, \bibinfo{author}{Aono\xfnm[ M]}, et~al.
\newblock \bibinfo{title}{Shrec’14 track: Extended large scale sketch-based
  3d shape retrieval}.
\newblock In: \bibinfo{booktitle}{Eurographics workshop on 3D object
  retrieval}; vol. \bibinfo{volume}{2014}. \bibinfo{year}{2014}{\natexlab{b}},
  p. \bibinfo{pages}{121--130}.
%Type = Inproceedings
\bibitem[{Li et~al.(2016)Li, Lu, Duan, Dong, Fan, Qian et~al.}]{li2016shrec}
\bibinfo{author}{Li\xfnm[ B]}, \bibinfo{author}{Lu\xfnm[ Y]},
  \bibinfo{author}{Duan\xfnm[ F]}, \bibinfo{author}{Dong\xfnm[ S]},
  \bibinfo{author}{Fan\xfnm[ Y]}, \bibinfo{author}{Qian\xfnm[ L]}, et~al.
\newblock \bibinfo{title}{Shrec'16 track: 3d sketch-based 3d shape retrieval}.
\newblock In: \bibinfo{booktitle}{Eurographics workshop on 3D object
  retrieval}. \bibinfo{year}{2016},.
%Type = Inproceedings
\bibitem[{Yuan et~al.(2018)Yuan, Li, Lu, Bai, Bai, Bui et~al.}]{yuan2018shrec}
\bibinfo{author}{Yuan\xfnm[ J]}, \bibinfo{author}{Li\xfnm[ B]},
  \bibinfo{author}{Lu\xfnm[ Y]}, \bibinfo{author}{Bai\xfnm[ S]},
  \bibinfo{author}{Bai\xfnm[ X]}, \bibinfo{author}{Bui\xfnm[ NM]}, et~al.
\newblock \bibinfo{title}{Shrec'18 track: 2d scene sketch-based 3d scene
  retrieval}.
\newblock In: \bibinfo{booktitle}{Eurographics Workshop on 3D Object
  Retrieval}. \bibinfo{year}{2018}, p. \bibinfo{pages}{29--36}.
%Type = Inproceedings
\bibitem[{Dey et~al.(2019)Dey, Riba, Dutta, Llados and Song}]{doodle}
\bibinfo{author}{Dey\xfnm[ S]}, \bibinfo{author}{Riba\xfnm[ P]},
  \bibinfo{author}{Dutta\xfnm[ A]}, \bibinfo{author}{Llados\xfnm[ J]},
  \bibinfo{author}{Song\xfnm[ YZ]}.
\newblock \bibinfo{title}{Doodle to search: Practical zero-shot sketch-based
  image retrieval}.
\newblock In: \bibinfo{booktitle}{Proceedings of the IEEE/CVF Conference on
  Computer Vision and Pattern Recognition}. \bibinfo{year}{2019}, p.
  \bibinfo{pages}{2179--2188}.
%Type = Article
\bibitem[{Sangkloy et~al.(2016)Sangkloy, Burnell, Ham and Hays}]{sketchy}
\bibinfo{author}{Sangkloy\xfnm[ P]}, \bibinfo{author}{Burnell\xfnm[ N]},
  \bibinfo{author}{Ham\xfnm[ C]}, \bibinfo{author}{Hays\xfnm[ J]}.
\newblock \bibinfo{title}{The sketchy database: learning to retrieve badly
  drawn bunnies}.
\newblock \bibinfo{journal}{ACM Transactions on Graphics}
  \bibinfo{year}{2016};\bibinfo{volume}{35}(\bibinfo{number}{4}):\bibinfo{pages}{1--12}.
%Type = Article
\bibitem[{Eitz et~al.(2012)Eitz, Hays and Alexa}]{tuberlin}
\bibinfo{author}{Eitz\xfnm[ M]}, \bibinfo{author}{Hays\xfnm[ J]},
  \bibinfo{author}{Alexa\xfnm[ M]}.
\newblock \bibinfo{title}{How do humans sketch objects?}
\newblock \bibinfo{journal}{ACM Transactions on graphics}
  \bibinfo{year}{2012};\bibinfo{volume}{31}(\bibinfo{number}{4}):\bibinfo{pages}{1--10}.
%Type = Inproceedings
\bibitem[{Ha and Eck(2018)}]{quickdraw}
\bibinfo{author}{Ha\xfnm[ D]}, \bibinfo{author}{Eck\xfnm[ D]}.
\newblock \bibinfo{title}{A neural representation of sketch drawings}.
\newblock In: \bibinfo{booktitle}{International Conference on Learning
  Representations}. \bibinfo{year}{2018},.
%Type = Inproceedings
\bibitem[{Uy et~al.(2019)Uy, Pham, Hua, Nguyen and Yeung}]{scanobjectnn}
\bibinfo{author}{Uy\xfnm[ MA]}, \bibinfo{author}{Pham\xfnm[ QH]},
  \bibinfo{author}{Hua\xfnm[ BS]}, \bibinfo{author}{Nguyen\xfnm[ T]},
  \bibinfo{author}{Yeung\xfnm[ SK]}.
\newblock \bibinfo{title}{Revisiting point cloud classification: A new
  benchmark dataset and classification model on real-world data}.
\newblock In: \bibinfo{booktitle}{Proceedings of the IEEE/CVF international
  conference on computer vision}. \bibinfo{year}{2019}, p.
  \bibinfo{pages}{1588--1597}.
%Type = Inproceedings
\bibitem[{Wu et~al.(2018)Wu, Xiong, Yu and Lin}]{modelnet40}
\bibinfo{author}{Wu\xfnm[ Z]}, \bibinfo{author}{Xiong\xfnm[ Y]},
  \bibinfo{author}{Yu\xfnm[ SX]}, \bibinfo{author}{Lin\xfnm[ D]}.
\newblock \bibinfo{title}{Unsupervised feature learning via non-parametric
  instance discrimination}.
\newblock In: \bibinfo{booktitle}{Proceedings of the IEEE conference on
  computer vision and pattern recognition}. \bibinfo{year}{2018}, p.
  \bibinfo{pages}{3733--3742}.
%Type = Article
\bibitem[{Chang et~al.(2015)Chang, Funkhouser, Guibas, Hanrahan, Huang, Li
  et~al.}]{shapenet}
\bibinfo{author}{Chang\xfnm[ AX]}, \bibinfo{author}{Funkhouser\xfnm[ T]},
  \bibinfo{author}{Guibas\xfnm[ L]}, \bibinfo{author}{Hanrahan\xfnm[ P]},
  \bibinfo{author}{Huang\xfnm[ Q]}, \bibinfo{author}{Li\xfnm[ Z]}, et~al.
\newblock \bibinfo{title}{Shapenet: An information-rich 3d model repository}.
\newblock \bibinfo{journal}{arXiv preprint arXiv:151203012}
  \bibinfo{year}{2015};.
%Type = Inproceedings
\bibitem[{Xiao et~al.(2010)Xiao, Hays, Ehinger, Oliva and
  Torralba}]{xiao2010sun}
\bibinfo{author}{Xiao\xfnm[ J]}, \bibinfo{author}{Hays\xfnm[ J]},
  \bibinfo{author}{Ehinger\xfnm[ KA]}, \bibinfo{author}{Oliva\xfnm[ A]},
  \bibinfo{author}{Torralba\xfnm[ A]}.
\newblock \bibinfo{title}{Sun database: Large-scale scene recognition from
  abbey to zoo}.
\newblock In: \bibinfo{booktitle}{2010 IEEE computer society conference on
  computer vision and pattern recognition}. \bibinfo{organization}{IEEE};
  \bibinfo{year}{2010}, p. \bibinfo{pages}{3485--3492}.
%Type = Inproceedings
\bibitem[{Su et~al.(2015)Su, Maji, Kalogerakis and
  Learned{-}Miller}]{su15mvcnn}
\bibinfo{author}{Su\xfnm[ H]}, \bibinfo{author}{Maji\xfnm[ S]},
  \bibinfo{author}{Kalogerakis\xfnm[ E]},
  \bibinfo{author}{Learned{-}Miller\xfnm[ EG]}.
\newblock \bibinfo{title}{Multi-view convolutional neural networks for 3d shape
  recognition}.
\newblock In: \bibinfo{booktitle}{Proc. ICCV}. \bibinfo{year}{2015},.
%Type = Inproceedings
\bibitem[{Simonyan and Zisserman(2015)}]{vgg}
\bibinfo{author}{Simonyan\xfnm[ K]}, \bibinfo{author}{Zisserman\xfnm[ A]}.
\newblock \bibinfo{title}{Very deep convolutional networks for large-scale
  image recognition}.
\newblock In: \bibinfo{booktitle}{International Conference on Learning
  Representations}. \bibinfo{year}{2015},.
%Type = Inproceedings
\bibitem[{Qi et~al.(2017)Qi, Su, Mo and Guibas}]{pointnet}
\bibinfo{author}{Qi\xfnm[ CR]}, \bibinfo{author}{Su\xfnm[ H]},
  \bibinfo{author}{Mo\xfnm[ K]}, \bibinfo{author}{Guibas\xfnm[ LJ]}.
\newblock \bibinfo{title}{Pointnet: Deep learning on point sets for 3d
  classification and segmentation}.
\newblock In: \bibinfo{booktitle}{Proceedings of the IEEE conference on
  computer vision and pattern recognition}. \bibinfo{year}{2017}, p.
  \bibinfo{pages}{652--660}.
%Type = Inproceedings
\bibitem[{He et~al.(2016)He, Zhang, Ren and Sun}]{resnet}
\bibinfo{author}{He\xfnm[ K]}, \bibinfo{author}{Zhang\xfnm[ X]},
  \bibinfo{author}{Ren\xfnm[ S]}, \bibinfo{author}{Sun\xfnm[ J]}.
\newblock \bibinfo{title}{Deep residual learning for image recognition}.
\newblock In: \bibinfo{booktitle}{Proceedings of the IEEE conference on
  computer vision and pattern recognition}. \bibinfo{year}{2016}, p.
  \bibinfo{pages}{770--778}.
%Type = Inproceedings
\bibitem[{Tan and Le(2021)}]{EfficientNetV2}
\bibinfo{author}{Tan\xfnm[ M]}, \bibinfo{author}{Le\xfnm[ QV]}.
\newblock \bibinfo{title}{Efficientnetv2: Smaller models and faster training}.
\newblock In: \bibinfo{editor}{Meila\xfnm[ M]}, \bibinfo{editor}{Zhang\xfnm[
  T]}, editors. \bibinfo{booktitle}{Proceedings of the 38th International
  Conference on Machine Learning, {ICML} 2021, 18-24 July 2021, Virtual Event};
  vol. \bibinfo{volume}{139} of \emph{\bibinfo{series}{Proceedings of Machine
  Learning Research}}. \bibinfo{publisher}{{PMLR}}; \bibinfo{year}{2021}, p.
  \bibinfo{pages}{10096--10106}.
%Type = Inproceedings
\bibitem[{Xu~Ma(2022)}]{pointMLP}
\bibinfo{author}{Xu~Ma Can~Qin\xfnm[ HYHRYF]}.
\newblock \bibinfo{title}{Rethinking network design and local geometry in point
  cloud: A simple residual {MLP} framework}.
\newblock In: \bibinfo{booktitle}{ICLR}. \bibinfo{year}{2022}, p.
  \bibinfo{pages}{1--15}.
%Type = Inproceedings
\bibitem[{Sung et~al.(2018)Sung, Yang, Zhang, Xiang, Torr and
  Hospedales}]{relation_net}
\bibinfo{author}{Sung\xfnm[ F]}, \bibinfo{author}{Yang\xfnm[ Y]},
  \bibinfo{author}{Zhang\xfnm[ L]}, \bibinfo{author}{Xiang\xfnm[ T]},
  \bibinfo{author}{Torr\xfnm[ PH]}, \bibinfo{author}{Hospedales\xfnm[ TM]}.
\newblock \bibinfo{title}{Learning to compare: Relation network for few-shot
  learning}.
\newblock In: \bibinfo{booktitle}{Proceedings of the IEEE conference on
  computer vision and pattern recognition}. \bibinfo{year}{2018}, p.
  \bibinfo{pages}{1199--1208}.
%Type = Article
\bibitem[{Lu et~al.(2018)Lu, Huang, Lin, Yang, Guo and Fu}]{2018Domain}
\bibinfo{author}{Lu\xfnm[ P]}, \bibinfo{author}{Huang\xfnm[ G]},
  \bibinfo{author}{Lin\xfnm[ H]}, \bibinfo{author}{Yang\xfnm[ W]},
  \bibinfo{author}{Guo\xfnm[ G]}, \bibinfo{author}{Fu\xfnm[ Y]}.
\newblock \bibinfo{title}{Domain-aware se network for sketch-based image
  retrieval with multiplicative euclidean margin softmax}
  \bibinfo{year}{2018};.
%Type = Article
\bibitem[{DeCarlo et~al.(2003)DeCarlo, Finkelstein, Rusinkiewicz and
  Santella}]{DeCarlo:2003:SCF}
\bibinfo{author}{DeCarlo\xfnm[ D]}, \bibinfo{author}{Finkelstein\xfnm[ A]},
  \bibinfo{author}{Rusinkiewicz\xfnm[ S]}, \bibinfo{author}{Santella\xfnm[ A]}.
\newblock \bibinfo{title}{Suggestive contours for conveying shape}.
\newblock \bibinfo{journal}{ACM Transactions on Graphics (Proc SIGGRAPH)}
  \bibinfo{year}{2003};\bibinfo{volume}{22}(\bibinfo{number}{3}):\bibinfo{pages}{848--855}.
%Type = Article
\bibitem[{Liu et~al.(2017)Liu, Shen, Shen, Liu and Shao}]{2017Deep}
\bibinfo{author}{Liu\xfnm[ L]}, \bibinfo{author}{Shen\xfnm[ F]},
  \bibinfo{author}{Shen\xfnm[ Y]}, \bibinfo{author}{Liu\xfnm[ X]},
  \bibinfo{author}{Shao\xfnm[ L]}.
\newblock \bibinfo{title}{Deep sketch hashing: Fast free-hand sketch-based
  image retrieval}.
\newblock \bibinfo{journal}{IEEE} \bibinfo{year}{2017};.
%Type = Article
\bibitem[{Jie et~al.(2017)Jie, Li, Gang and Albanie}]{2017Squeeze}
\bibinfo{author}{Jie\xfnm[ H]}, \bibinfo{author}{Li\xfnm[ S]},
  \bibinfo{author}{Gang\xfnm[ S]}, \bibinfo{author}{Albanie\xfnm[ S]}.
\newblock \bibinfo{title}{Squeeze-and-excitation networks}.
\newblock \bibinfo{journal}{IEEE Transactions on Pattern Analysis and Machine
  Intelligence}
  \bibinfo{year}{2017};\bibinfo{volume}{PP}(\bibinfo{number}{99}).

\end{thebibliography}

\end{document}